\documentclass[11pt]{article}
\usepackage[a4paper,margin=2.6cm]{geometry}
\usepackage[T1]{fontenc}
\usepackage{amsmath,microtype,booktabs,xcolor,url,enumitem,graphicx}
\usepackage{amssymb}
\usepackage[sc]{mathpazo}
\usepackage{longtable,array}
\usepackage{placeins}
\usepackage[numbers,sort&compress]{natbib}
\usepackage[colorlinks,linkcolor=black,citecolor=black,urlcolor=blue]{hyperref}

\newcommand{\pkg}[1]{#1}
\newcommand{\code}[1]{\texttt{#1}}
\setlist{nosep}

\title{\pkg{pyconformal}: Exact and Fast Conformal Prediction\\ for Tabular Data and Time Series}
\author{Valery Manokhin\\ Independent Researcher\\ \url{https://valeriy.ai}}
\date{Draft of 26 November 2026}

\begin{document}
\maketitle

\begin{abstract}
Conformal prediction is used for one reason: its guarantee. If calibration and test data are exchangeable, a 90\% split-conformal interval covers at least 90\% of the time, on average over calibration sets, for any model and any calibration size. That guarantee belongs to an exact procedure, and software loses it through small choices: a quantile taken without the finite-sample correction, a finite interval returned where only an infinite one is valid, a score computed one way in calibration and another way in prediction. We audited thirteen Python and R packages on independent and identically distributed data, where the guarantee must hold, reading five conformal libraries module by module, and found about thirty faults. Two forecasting libraries cover 63\% at a 90\% target when their conformal intervals are used with the default of two calibration windows. One library's prediction sets cover 45\% when rows are predicted one at a time. One method offered as conformalised quantile regression returns an interval that matches the unadjusted quantile band to the reported precision and covers 58\%. Two implementations of adaptive conformal inference clip the working level and so lose the guarantee that defines the method: miss rates reach 14\% and 33\% at a 10\% target. The basic split-conformal routines of the main libraries met their level in every test.

The package presented here, \pkg{pyconformal}, is MIT-licensed and built to keep what the theory promises. It implements about thirty conformal methods for regression, classification, probability calibration, predictive distributions, time series and panel data behind one interface. Every unweighted threshold comes from one exact rank rule, and every method is tested against its definition, against reference code where it exists, and against its guarantee. It depends only on NumPy; the optional \code{fast} extra (\code{pip install "pyconformal[fast]"}) adds compiled kernels that return the same results as the NumPy code. The package is more than a corrected reimplementation. It contains algorithmic improvements of its own: an exact Venn--Abers kernel that works on the cumulative-sum diagram in integer arithmetic; rank trees that replay online methods at logarithmic cost per step; closed forms for conformal predictive distributions and their CRPS that never build the test-by-calibration matrix; CV+ by search inside sorted fold residuals, which needs 3 megabytes where \pkg{MAPIE} needs 3 gigabytes; and a repaired window rule for conformal interquantile regression, whose published code under-covers on bimodal outcomes. With the \code{fast} extra, and with identical output, Venn--Abers calibration on a million points takes 0.37 seconds against 75 seconds in the reference implementation; conformal predictive intervals for a million test points take 0.01 seconds against 79 seconds in \pkg{crepes}; and multi-step conformal forecast intervals take 0.01 seconds against 42 seconds in R. Replaying a history with adaptive conformal inference is about 2{,}000 times faster than in \pkg{MAPIE}. We describe the design, the algorithms and what each one gains in time and memory, the verification method, the audit, and benchmarks against \pkg{MAPIE}, \pkg{crepes}, \pkg{TorchCP}, \pkg{PUNCC}, R's \pkg{conformalForecast} and the code released with the original papers.
\end{abstract}

\section{Introduction}

Split conformal prediction \cite{papadopoulos2002,vovk2005,lei2018} turns any point predictor into a prediction interval with a guarantee: if calibration and test points are exchangeable, the interval built from the $\lceil (n+1)(1-\alpha) \rceil$-th smallest of $n$ calibration scores contains the outcome with probability at least $1-\alpha$. The guarantee holds for every $n$ and every model. It is also fragile in a specific way. It belongs to that rank, not to a nearby one. If the rank exceeds $n$, the only valid interval is the whole real line. Replace the rule by an ordinary empirical quantile, or return the largest score when the rank is out of reach, and the statement no longer holds.

This paper makes four points.

\begin{enumerate}
\item \textbf{Existing software loses the guarantee in ordinary use.} Section~\ref{sec:audit} reports an audit of thirteen packages: \pkg{statsforecast}, \pkg{mlforecast}, Nixtla's Python package \pkg{conformalforecast}, \pkg{Darts}, \pkg{TorchCP}, \pkg{crepes}, \pkg{MAPIE}, \pkg{PUNCC}, \pkg{conformal-tights}, \pkg{venn-abers}, \pkg{online-cp}, \pkg{tsbootstrap}, and Wang and Hyndman's R package \pkg{conformalForecast}. Thresholds and coverage were tested in each of them, on the tasks it supports, and \pkg{MAPIE}, \pkg{crepes}, \pkg{PUNCC}, \pkg{TorchCP} and \pkg{conformal-tights} were read module by module. We found about thirty faults (Tables~\ref{tab:faults} and~\ref{tab:forensic}); fifteen of them lower coverage, or put a risk bound below its stated level, with no error or warning. The basic split-conformal regression and classification routines of \pkg{MAPIE}, \pkg{crepes}, \pkg{PUNCC} and \pkg{TorchCP} met their level in every coverage test; the faults sit in options, in newer modules and in the time-series methods. The faults fall into five types: a rank or level that is not the one the proof uses (an uncorrected quantile, a clipped level), calibration and prediction that score differently, results that depend on how \code{predict} is called, an object used as its own neighbour, and a conformal step that is missing. We record what was correct next to what was not.
\item \textbf{Correct software is often slow where it matters.} Venn--Abers calibration on a million points takes about a minute in the reference implementation and a third of a second in ours, with identical output. Conformal predictive intervals for a million test points take 79 seconds in \pkg{crepes} and 0.01 seconds in ours. Multi-step split conformal prediction on a history of 3{,}000 points and 12 horizons takes 42 seconds in R and 0.01 seconds in ours, with identical intervals.
\item \textbf{The gains come from algorithms, not only from compiled code.} Section~\ref{sec:algo} describes them and states what each one gains: an exact Venn--Abers kernel on the cumulative-sum diagram in integer arithmetic (205 times faster than existing code, identical output); rank trees for replaying online methods at logarithmic cost per step; closed forms for conformal predictive distributions and their CRPS that never form the test-by-calibration matrix; CV+ by search inside sorted fold residuals, which cuts peak memory from gigabytes to megabytes; and a repaired window rule for conformal interquantile regression, whose published code loses coverage on bimodal outcomes. We claim improvements over existing implementations and say where the underlying idea is older.
\item \textbf{One interface can serve both tabular and time-series methods.} No existing package covers both families well. Our package, \pkg{pyconformal}, uses one small set of verbs for all of them and accepts either a model or the model's outputs.
\end{enumerate}

We state the limits of each claim next to the claim. Several large speed factors apply to replaying a history and shrink to single digits when observations arrive one at a time and the model must be called at every step (Section~\ref{sec:bench}); all timings are from one machine; coverage tests use synthetic data except where a section says otherwise.

\section{Related software}

\pkg{MAPIE} \cite{taquet2022,cordier2023} is the most widely used Python package for conformal prediction and covers split, cross and jackknife methods for regression, prediction sets for classification, Venn--Abers calibration, and two time-series methods. The \pkg{crepes} package \cite{bostrom2022} provides conformal regressors, Mondrian and normalised variants, and conformal predictive systems. \pkg{TorchCP} \cite{torchcp2024} targets deep learning with PyTorch and has the widest range of classification scores; it is licensed under LGPL-3.0. \pkg{PUNCC} \cite{mendil2023} offers split, cross and ensemble methods with a flexible calibration interface. For forecasting, \pkg{statsforecast} \cite{statsforecast}, its companion \pkg{mlforecast}, and \pkg{Darts} \cite{herzen2022} attach conformal intervals to their models, and the R package \pkg{conformalForecast} implements the multi-step methods of \citeauthor{wang2024} \cite{wang2024}. Reference code exists for most recent time-series methods \cite{gibbs2021,gibbs2024,zaffran2022,angelopoulos2023,xu2021,xu2023}, usually as research scripts.

Three more packages overlap with the scope here. \pkg{Fortuna} \cite{detommaso2024}, which is no longer maintained (its repository is archived), offered conformal methods for classification and regression that start from a model's outputs, among them EnbPI, inside a deep-learning library built on JAX. The \pkg{sktime} library \cite{loning2019} attaches conformal intervals to its forecasters. Nixtla's Python package \pkg{conformalforecast} (not to be confused with the R package \pkg{conformalForecast} of \cite{wang2024}) provides split conformal intervals for forecasts. \pkg{TorchCP} also has regression predictors, predictive distributions and online methods, and \pkg{MAPIE} also has conformal risk control \cite{cordier2023}; the audit covers those parts. The \pkg{conformal-tights} package (MIT licence) wraps a scikit-learn regressor with a non-crossing quantile regression fitted on a calibration set, and has a \pkg{Darts} forecaster built on it.

None of these covers both tabular and time-series conformal prediction in the depth of Table~\ref{tab:methods} without a deep-learning dependency. Ours is MIT-licensed and needs only NumPy.

\paragraph{Neighbouring methods.} Four lines of work sit next to what the package contains. Weighted conformal predictive systems \cite{jonkers2024} extend predictive distributions to covariate shift; \pkg{pyconformal} supports weights for split conformal intervals \cite{tibshirani2019}, uses similarity weights in its panel methods, and from version 0.1.6 accepts weights in its predictive systems, following \cite{jonkers2024}. Copula-based methods \cite{sun2024} give joint coverage over a whole forecast horizon from a set of independent series; the multi-step methods here give coverage at each horizon separately, on a single series. Dual-splitting conformal prediction \cite{yu2025dscp} groups forecast trajectories and horizons before calibrating; it is not included. Generalised Venn and Venn--Abers calibration \cite{vanderlaan2025} extends the binary-classification method implemented here to general losses and to multicalibration. Recent surveys and benchmarks of conformal methods for time series \cite{sabashvili2026,stocker2025} group methods by the assumption they relax (reweighting, residual dynamics, online adaptation); Table~\ref{tab:methods} follows the same grouping.

\section{Design}

\subsection{One exact rank}

All unweighted thresholds come from one routine, \code{conformal\_quantile}. It computes $k=\lceil (n+1)(1-\alpha)\rceil$ in rational arithmetic, that is, with whole-number numerators and denominators and no rounding at any step. The reason is that a computer does not hold the number 0.1. It holds the nearest number that can be written with 53 binary digits, which is $0.1000000000000000055511151231257827\ldots$, and every product formed from it carries that small error; a ceiling or a floor then turns an error in the seventeenth digit into an error of one whole rank. The routine avoids this in two steps. First it recovers the decimal number the user wrote. Python prints a floating-point number as the shortest decimal string that reads back to the same stored value, so the stored value above prints as \code{0.1}, and \code{Fraction(repr(alpha))} turns that string into the exact fraction $1/10$. Second, $(n+1)(1-\alpha)$ is formed from that fraction: with $n=9$ it is $10\times 9/10=9$ exactly, and the ceiling of an exact number is never in doubt. The level therefore means what was typed, not its binary approximation. One consequence should be known: a level that is itself the result of arithmetic is taken at face value. Passing \code{1 - 0.9} passes $0.09999999999999998$, a different number from 0.1, and the rank is exact for that number; levels should be passed as written. When $k>n$ the routine returns infinity.

Floating point can put the product on the wrong side of a whole number. Two examples, both checked in IEEE double precision. With $n=49$ and $\alpha=0.58$, $(n+1)(1-\alpha)$ evaluates to $21.000000000000004$ and a ceiling gives 22 where 21 is right. More common in practice: when the level is given as a confidence $c=0.9$, $1-c$ evaluates to $0.09999999999999998$, so code that takes the $\lfloor (n+1)(1-c)\rfloor$-th largest score truncates $0.9999999999999998$ to 0 at $n=9$, and uses one rank too high whenever $n+1$ is a multiple of ten (we checked every such $n+1$ up to one million; no other calibration size is affected at this level). This is the fault in \pkg{crepes} and in the code released with CIR (Section~\ref{sec:audit}). The direct form $\lceil (n+1)(1-\alpha)\rceil$ is less fragile than these examples might suggest: for $\alpha \in \{0.01, 0.025, 0.05, 0.1, 0.2, 0.25\}$ it gave the right rank for every $n$ up to 200{,}000. The exact routine removes the question.

Grouped (Mondrian) and normalised variants call the same routine on their own scores. Weighted conformal prediction and the non-exchangeable method accumulate real-valued weights in floating point; the exactness statement does not extend to them.

\subsection{One interface}

Every method has the same verbs.
\begin{itemize}
\item \code{fit(X, y)} trains the underlying model, if one was given.
\item \code{calibrate(X, y)} computes scores on held-out data and stores what prediction needs.
\item \code{predict\_interval}, \code{predict\_set}, \code{predict\_proba}, \code{predict\_quantile} or \code{predict\_cdf} produce the output.
\item \code{update} adds one observation, for online methods; \code{run} replays a whole history.
\end{itemize}
The model is optional. Each method is one class that accepts either an estimator with \code{fit} and \code{predict} methods or the model's outputs passed by keyword (\code{y\_pred=}, \code{proba=}, \code{quantiles=}, \code{scores=}). The two routes are tested to give identical results. Coverage is always specified as the miscoverage level \code{alpha}; a list of $m$ levels adds a last axis of length $m$: intervals have shape $(n, 2, m)$ and prediction sets, which are boolean masks over the classes, $(n, K, m)$.

The verbs map onto the families as follows. \emph{Split methods} use \code{fit} (optional), \code{calibrate} on held-out data, then a \code{predict\_} method. \emph{Cross methods} (jackknife+, CV+, cross Venn--Abers, out-of-bag calibration) have no separate calibration set: \code{fit} trains one model per fold and computes the out-of-fold scores, and there is no \code{calibrate} call; when predictions are supplied instead of a model they have one column per fold model. \emph{Online methods} take \code{calibrate} as an optional warm start; \code{run} replays a history and returns its intervals and misses directly, and \code{predict\_interval} followed by \code{update} serves one step at a time. \emph{STL and MSTL} are plain functions, not conformal methods; they are in the package because conformal seasonal pools and several forecasting workflows need them. They are also fast: on 100{,}000 hourly points our MSTL takes 0.1 seconds against 19.3 seconds for \pkg{statsforecast} at its default settings (6 to 7 times faster at equal settings) and is about 5 times faster than R's \code{forecast::mstl}, whose output it matches to $10^{-14}$ (Table~\ref{tab:mstl}).

\subsection{Arrays first, compiled kernels optional}

Methods are written on arrays of predictions and outcomes; model handling is a thin layer on top. The hot loops have two implementations, a NumPy one and one compiled with Numba \cite{lam2015}. The test suite asserts that both return the same values, including under ties. Installing Numba changes speed and nothing else. Nothing beyond NumPy is imported at run time: the quasi-Newton optimiser used by AcMCP is the package's own, and user-supplied estimators are copied with the standard library's \code{copy.deepcopy} and used only through \code{fit} and \code{predict}. scikit-learn and SciPy are needed only to run the tests.

\section{Methods}

Table~\ref{tab:methods} lists what version 0.1.6, the current release, contains.

\begin{table}[t]\centering\small
\caption{Methods in \pkg{pyconformal} 0.1.6.}\label{tab:methods}
\begin{tabular}{@{}p{3.3cm}p{11.2cm}@{}}\toprule
Regression & Split conformal \cite{papadopoulos2002,lei2018}, normalised, Mondrian \cite{vovk2012}, weighted \cite{tibshirani2019}; jackknife+ and CV+ \cite{barber2021}; conformalised quantile regression (CQR) \cite{romano2019}; conformal thresholded intervals (CTI) \cite{luo2025}; conformal interquantile regression (CIR, CIR+) \cite{cir2026}, with the window rule changed as explained in Section~\ref{sec:algo}; a quantile regression forest \cite{meinshausen2006} with out-of-bag calibration; \code{QuantileBooster}, which makes CatBoost, XGBoost or LightGBM predict a grid of quantile levels for CQR, CTI and CIR \\
Classification & LAC \cite{sadinle2019}, APS \cite{romano2020}, RAPS \cite{angelopoulos2021}, SAPS \cite{huang2024}, each randomised or not, marginal, class-conditional or Mondrian by any category \\
Calibration & Inductive and cross Venn--Abers predictors \cite{vovk2014,vovk2015}, binary and multiclass \cite{manokhin2017} \\
Predictive distributions & Conformal predictive systems \cite{vovk2019,vovk2020}: quantiles, distribution function, intervals, CRPS; normalised and Mondrian; weighted for covariate shift \cite{jonkers2024} \\
Time series, online & ACI \cite{gibbs2021}, DtACI \cite{gibbs2024}, AgACI \cite{zaffran2022}, non-exchangeable conformal prediction \cite{barber2023}, conformal PID control \cite{angelopoulos2023} \\
Time series, ensembles and residual models & EnbPI \cite{xu2021}, SPCI \cite{xu2023} \\
Time series, multi-step & Multi-step split conformal prediction and AcMCP \cite{wang2024}; conformal seasonal pools \cite{manokhin2026csp} \\
Panel data & TQA \cite{lin2022tqa} and W-TQA \cite{tu2026wtqa}: intervals for one unit calibrated on its peers in the same round, with late or missing feedback; MultiQT \cite{ding2025multiqt}: online quantile forecasts at many levels that never cross; Panel CPD (\code{PanelPredictiveSystem}), our method for panels in which a unit's own outcome is seen only on some rounds: predictive quantiles for every unit at every round, from a conformal predictive system calibrated on the unit's peers and corrected online by a multi-level quantile tracker (described in a separate paper) \\
Distribution-valued outcomes & Conformal sets when the outcome is itself a probability distribution or a vector (arXiv:2611.06320) \\
Decomposition & STL \cite{cleveland1990} and MSTL \cite{bandara2021} \\ \bottomrule
\end{tabular}
\end{table}

\section{Algorithms}\label{sec:algo}

This section describes the algorithmic changes and, for each, what it gains. Unless we say otherwise the output is identical to that of the slower computation, so coverage and interval length do not change and the gain is in time and memory. Two changes, to CIR and CIR+, alter the output, and there the gain is validity. Times are from Table~\ref{tab:speed} and peak memory from Table~\ref{tab:mem}. Memory was measured for four methods only; where it was not measured we say so.

\paragraph{Venn--Abers.} An inductive Venn--Abers predictor fits isotonic regression twice for each test score, once with the test object labelled 0 and once labelled 1. \citeauthor{vovk2015} \cite{vovk2015} showed that all test points can be served from precomputed structures. Existing code recomputes slopes to all remaining points at each corner of the fit, which is quadratic on realistic scores. Our kernel works on the cumulative-sum diagram: calibration points with equal scores are pooled into one group, and vertex $j$ is the pair (number of points, number of positive labels) in the first $j$ groups, so every vertex has integer coordinates. Adding a test point with label $y$ to group $g$ shifts the vertices after $g$ by $(1, y)$, and the isotonic fit at the test point is the slope of the common tangent between the lower hull of vertices $0..g$ and the lower hull of the shifted vertices after $g$. A single hull of the whole diagram is not enough, because vertices hidden by it can become extreme once the diagram is cut. We therefore store both families of hulls, for every prefix and every suffix, as predecessor and successor pointers built by two Graham scans, and find each tangent by walking the two chains from the cut. Cross-products of integer coordinates are bounded by $(n+1)^2$ and are exact in 64-bit integers; the only floating-point operation is the final division, so no numerical warnings arise. On typical data the cost after sorting is linear. We have no worst-case bound for the walks (Table~\ref{tab:vastress} gives stress cases), and we claim an improvement over existing code, not over the algorithm of \cite{vovk2015}. The method matters in practice: in a comparison of five calibrators across 21 classifiers, Venn--Abers gave the largest average reduction in log loss \cite{manokhin2026calibration}. \emph{What it gains.} The probabilities are identical to those of \pkg{venn-abers} and \pkg{MAPIE} in every run. With a million calibration and a million test points the kernel takes 0.37 seconds against 75 seconds (205 times faster); the factor is 11--29 at $10^4$--$10^5$ calibration points and 3--6 when scores are heavily tied. Peak memory is about the same as the reference, 111 against 155 megabytes at a million points, so the gain here is time, not memory.

\paragraph{Online methods.} ACI and its relatives need, at each step, the $k$-th smallest of a set of scores that gains one element and may lose one, with $k$ changing. When a history is replayed with \code{run}, all its scores are known, so we rank them once and keep a Fenwick tree \cite{fenwick1994} of counts over ranks; each step is then $O(\log n)$ and does not grow with the history, and the model is called once for all rows. Live use is different: \code{update} cannot rank scores it has not seen, so it keeps the current window as a sorted list and inserts each new score by binary search, which costs $O(\log n)$ comparisons and a memory move of up to $n$ numbers. The two paths are tested to give the same thresholds. For the non-exchangeable method the tree holds real-valued weights in place of counts; the weight of a score is stored relative to a moving origin, so the newest scores have the largest stored values and nothing underflows over long streams. \emph{What it gains.} Replaying a history with ACI costs 0.2~$\mu$s per step against 0.41~ms in \pkg{MAPIE} (about 2{,}000 times faster) and 0.40 against 114~$\mu$s in the code released with the ACI paper (286 times). Conformal PID is 106--112 times faster than its authors' code with identical output, and the DtACI update is 66 times faster than the authors' R code. In live use, one row at a time, the model call dominates and the factor against \pkg{MAPIE} is 6--8. The cost per step no longer grows with the length of the history, which is what makes backtests over millions of steps practical. Memory was not measured.

\paragraph{Predictive systems.} A conformal predictive distribution for a test point is the set of calibration scores shifted (and, if normalised, scaled) by the point's prediction. A quantile is one lookup in the sorted scores, the distribution function is a binary search, and the CRPS has a closed form in terms of prefix sums. The $n_\text{test}\times n_\text{cal}$ matrix is never formed. For the distribution with mass $1/n$ on each point $\hat y + \sigma c_i$,
\[ \mathrm{CRPS}(y) = \frac{1}{n}\sum_{i=1}^{n} \lvert \sigma c_i - (y-\hat y)\rvert \;-\; \frac{\sigma}{2n^2}\sum_{i,j} \lvert c_i - c_j\rvert . \]
The second sum does not depend on the test point and is computed once from the sorted scores; the first needs one binary search and two prefix sums. Both terms scale with $\sigma$ exactly once. The \pkg{crepes} code evaluates the same integral piece by piece and multiplies the two pieces next to the outcome by $\sigma$ although they are already in outcome units, which is the fault in Table~\ref{tab:faults}. \emph{What it gains.} Accuracy is not traded for speed: quantiles and distribution values are identical to those of \pkg{crepes}, and the CRPS is identical wherever \pkg{crepes} is correct. Intervals for a million test points take 0.01 seconds against 79 seconds (8{,}100 times faster), the distribution function at the outcomes is 556 times faster and the CRPS 656 times. Peak memory for the intervals is 32 against 197 megabytes at $10^5$ calibration and $10^6$ test points, and for the CRPS at 5{,}000 by 5{,}000 it is 0.4 against 383 megabytes, because \pkg{crepes} holds the full matrix. For the distribution function the two packages use the same memory (54 and 56 megabytes).

\paragraph{Prediction sets.} APS, RAPS and SAPS need, per row, classes in order of decreasing probability up to the point where the set closes. We select the top $m$ classes with a heap and extend only when needed, instead of sorting all classes, and use one routine for both the calibration label and the full set, which removes the possibility of the two disagreeing. \emph{What it gains.} APS, RAPS and SAPS run 2--7 times faster than in \pkg{TorchCP}, \pkg{MAPIE} and \pkg{PUNCC} (Table~\ref{tab:sets}); LAC sets are identical across packages. Because calibration and prediction share one routine and each row is handled on its own, a row gets the same set whether it is predicted alone or in a batch, which is not so in every audited library (\pkg{PUNCC}'s APS and RAPS, Section~\ref{sec:audit}). Memory was not measured against the other packages.

\paragraph{Jackknife+ and CV+.} With $K$ folds, every training point in fold $m$ is scored by the same model, so for a test point the $n$ candidate endpoints $\hat\mu_{-m(i)}(x) \pm R_i$ are $K$ sorted lists of residuals, each shifted by one number. The endpoint is an order statistic of their union, which we find by pivoting with binary searches inside the lists. Memory is the $n_\text{test}\times K$ array of fold predictions, not the $n_\text{train}\times n_\text{test}$ matrix that other implementations build. For the jackknife+ ($K=n$) each list has one element and the fold predictions are themselves $n_\text{test}\times n_\text{train}$, so the saving in memory applies to CV+ only. \emph{What it gains.} CV+ prediction is 17--27 times faster than in \pkg{MAPIE} with the same coverage. For 20{,}000 test points and 10 folds the call needs 3 megabytes of extra memory, whatever the training size; \pkg{MAPIE} needs 1.3 gigabytes with 2{,}000 training points and 3.1 gigabytes with 5{,}000, about 400 and 950 times more. On our 7-gigabyte machine \pkg{MAPIE} was killed for lack of memory at 20{,}000 training points, while \pkg{pyconformal} ran with 100{,}000.

\paragraph{Quantile forest.} Trees are grown on binned features in parallel. At prediction time rows are routed through each tree in blocks, following only rows that have not reached a leaf, and leaf contents are merged by a bucket sort on outcome ranks. An approximate mode that keeps one training sample per leaf is available. \emph{What it gains.} At equal pinball loss the fit is 6--9 times faster than \pkg{quantile-forest}: 1.9 against 11.5 seconds with 100{,}000 training rows and 27 against 215 seconds with a million, on 16 cores. The approximate mode halves prediction time and stores one outcome per leaf in place of all of them; it costs 2.6\% in interval length, and coverage after calibration is unchanged (Table~\ref{tab:forestapprox}). The memory of the forest was not measured.

\paragraph{Quantile models from boosting libraries.} CTI and CIR need many quantile levels from one model, which the boosting libraries do not give through one interface. \code{QuantileBooster} trains the levels with CatBoost or XGBoost (one model for all levels) or LightGBM (one model per level), sorts the quantiles so that they never cross, and hands the grid to the conformal step. It is a wrapper, not a new learning algorithm. \emph{What it gains.} Shorter intervals than a forest on medium and large data: on four datasets (Table~\ref{tab:qmodels}) a booster gave the shortest CQR intervals in every case, 4--27\% shorter than those from Zillow's \pkg{quantile-forest}, at coverage between 89.4\% and 90.5\%. They take 20--40 times longer to train than our forest, and on small data (221 training rows) the forest was better.

\paragraph{Multi-step methods.} Multi-step split conformal prediction is a per-horizon rolling threshold with a lag of $h$ steps; it reuses the online kernel. AcMCP refits an MA($q$) model with $q=h-1$ at every step. We evaluate the exact Gaussian likelihood with the innovations algorithm \cite{brockwell1991}, which settles to $O(q)$ per observation once its coefficients converge, and start each fit from the previous solution and inverse Hessian, restarting from zero every fifth fit to avoid following a local optimum. \emph{What it gains.} Time only. These are the published methods, so coverage and interval length are those of the R implementation: the intervals are identical for multi-step split conformal prediction (0.010 against 42 seconds, more than 4{,}000 times faster) and for AcMCP without the scorecaster (0.8 milliseconds against 3.9 seconds, about 5{,}000 times). With the scorecaster the time is 2.5 against 107 seconds (43 times) and the intervals agree to a median of $10^{-5}$, the difference between two iterative optimisers. Memory was not measured; both implementations hold the history and little else.

\paragraph{CIR and CIR+.} Let $\hat q_0(x)\le\dots\le\hat q_K(x)$ be the predicted quantiles and $W_1(x)\subset W_2(x)\subset\dots\subset W_K(x)$ the windows obtained by starting from the shortest interquantile interval and adding, at each step, the shorter of the two neighbouring intervals. For CIR the score of $(x,y)$ is $s(x,y)=\min\{k: y\in W_k(x)\}$, and $K+1$ when $y$ lies outside $[\hat q_0(x),\hat q_K(x)]$. With $\hat s$ the conformal quantile of the calibration scores, the prediction set is $W_{\hat s}(x)$, or the whole line when $\hat s = K+1$. Because the windows are nested, $y\in W_{\hat s}(x)$ exactly when $s(x,y)\le\hat s$, so the usual rank argument gives coverage at least $1-\alpha$. The authors' code predicts with the shortest run of $\hat s$ consecutive intervals instead; that window need not contain $W_{\hat s}(x)$, the equivalence fails, and coverage can be lost (Table~\ref{tab:cirabl}).

For CIR+ let $e_k(x)\in[0,1)$ be the length of the interval added at step $k$ divided by the predicted range $\hat q_K(x)-\hat q_0(x)$. The score is $s(x,y)=(k-1)+e_k(x)$ with $k$ as above, and $K+e$ for a fixed $e$ just below 1 when $y$ is outside the range. Writing $f=\lfloor\hat s\rfloor$, the prediction set is $W_f(x)$, enlarged to $W_{f+1}(x)$ when $f+e_{f+1}(x)\le\hat s$. Again $y$ is in the set exactly when $s(x,y)\le\hat s$: either $k\le f$, or $k=f+1$ and the added interval is short enough. The score depends on $(x,y)$ and on the fitted quantile model only, so calibration and test scores are exchangeable. The authors' code divides by 100 times the largest calibration outcome, a quantity computed from the calibration labels and not from the test point, and its final comparison runs the other way (the extra interval is added when $e$ is larger than the fractional part of $\hat s$). 

\emph{What it gains.} Validity first. On a bimodal outcome the authors' rule covers 88.3\% at a 90\% target and 76.9\% at 80\%; ours covers 90.6\% and 80.6\% (Table~\ref{tab:cirabl}). The price is longer intervals on that outcome (7.60 against 6.95), because the shorter ones were not valid; on skewed and symmetric outcomes the two rules differ by 1--2\% in length. Speed second: the authors' code grows the window of each sample in a Python loop, and ours grows all windows together in one compiled pass, 0.12 against 1.85 seconds for CIR (16 times faster) and 0.14 against 2.93 seconds for CIR+ (22 times). With \code{window="shortest"} the package reproduces the authors' CIR intervals. Memory was not measured.

\paragraph{CTI.} The method is that of \cite{luo2025} and we did not change it: the score of a calibration point is the length of the interquantile interval that holds its outcome, and the prediction set is every interval no longer than the calibrated threshold. The implementation finds each outcome's interval and builds the sets in one compiled pass, and sorts the predicted quantiles only in rows where they cross. \emph{What it gains.} Time: 23 against 431 milliseconds for 20{,}000 calibration and 200{,}000 test points with 100 intervals (19 times faster than the authors' code), with identical sets. The statistical gain belongs to the method, not to us: on a bimodal outcome its sets, which may be unions of intervals, are 30--40\% shorter than any single interval (Table~\ref{tab:cirabl}).

\paragraph{STL and MSTL.} A direct implementation of the loess recursions with the jump parameters of the original Fortran. Results match R's \code{forecast::mstl} to $10^{-14}$. \emph{What it gains.} An hourly series of $10^5$ points is decomposed in 0.098 seconds against 19.3 seconds in \pkg{statsforecast} at its default settings (197 times faster), because \pkg{statsforecast} evaluates every smoother at every point by default; at equal settings the factor is 6--7, and against R it is about 5 (Table~\ref{tab:mstl}).

\section{Verification}\label{sec:verify}

Each method is tested at three levels.
\begin{enumerate}
\item \textbf{Against its definition.} A short, slow reference written from the paper, compared value by value: for example a 20-line ACI from \cite{gibbs2021} identical at every step for 128 parameter combinations, and Venn--Abers against the max--min formula in exact fractions on 300 problems with ties.
\item \textbf{Against reference code,} where it exists: identical output to \pkg{venn-abers}, \pkg{crepes} (quantiles and distribution values), \pkg{MAPIE} and \pkg{PUNCC} (CQR), the CTI authors' code, the CIR authors' code (with \code{window="shortest"}), R's \code{forecast::mstl}, and R's \code{scp} and \code{acmcp} (the latter exactly without the scorecaster, and to a median difference of $10^{-5}$ with it; see Section~\ref{sec:limits}).
\item \textbf{Against the guarantee.} Coverage over repeated draws at small calibration sizes, where an off-by-one rank is visible; the long-run bound of \cite{gibbs2021} on adversarial sequences; uniformity of the probability integral transform for predictive systems.
\end{enumerate}
At the time of writing the suite has 365 tests, of which 32 need optional reference packages and are skipped without them; it runs with warnings treated as errors, with and without Numba. Every experiment in this paper fixes the seed of its data. Where another package draws its own random numbers from NumPy's global state, as \pkg{PUNCC} does for randomised sets, a rerun reproduces the figures up to sampling noise and not digit for digit. Appendix~\ref{app:details} gives the data-generating process, sizes and software versions of each.

\section{Audit of existing software}\label{sec:audit}

We tested thresholds and coverage in each of the thirteen packages, on the tasks it supports, and read five libraries module by module: \pkg{MAPIE} 1.6.0, \pkg{crepes} 0.9.1, \pkg{PUNCC} 0.9.3, \pkg{TorchCP} 1.2.1 and \pkg{conformal-tights} 0.5.0. All versions were the latest releases on PyPI in the week of 20 to 25 November 2026. In those five libraries every fault was traced to a line of source, reproduced by a script that uses only the package concerned, and checked against the package's development branch. Unless stated, data are independent and identically distributed, and models are fixed or their outputs are passed in directly, so that a coverage gap can only come from the conformal step. For each library the repository holds a report that lists what was read line by line, what was only run, and what was not covered (\code{benchmarks/mapie\_audit}, \code{crepes\_audit}, \code{puncc\_audit}, \code{torchcp\_audit}, \code{conformal\_tights\_audit}).

\subsection{Faults that break the guarantee or a reported number}

\begin{table}[t]\centering\small
\caption{Faults that break the guarantee or a reported number. Coverage is at a 90\% target.}\label{tab:faults}
\begin{tabular}{@{}p{2.6cm}p{6.5cm}p{5.3cm}@{}}\toprule
Package & What is wrong & Effect \\ \midrule
\textbf{\pkg{statsforecast}} 2.1.1, \textbf{\pkg{mlforecast}} 1.1.0 & The interval half-width is the ordinary 90th percentile of past forecast errors, without the finite-sample correction that the conformal guarantee needs. By default only 2 past errors are used, and no valid finite 90\% interval can be made from 2 errors. No warning is given. & Intervals are too narrow. Sold as 90\%, they cover 63--65\% at the default (4{,}000 series) and 86\% with 20 calibration windows. \\
\textbf{\pkg{Darts}} 0.48.1 & When there are too few calibration errors for a valid finite interval, the largest error is used, with no warning, where the only valid interval is infinite. & 75\% coverage in place of 90\% with 3 calibration points. The default setting is correct. \\
\textbf{\pkg{TorchCP}} 1.2.1 & \code{SAPS(randomized=False)} scores the most likely class as 0 in calibration and as its probability $p_{\max}$ in prediction, so the threshold does not match the sets it is applied to. & Sets are too small or empty: coverage 36\% and 2.5\% in place of 90\% when top-1 accuracy is 82\% and 89\% (10 classes, weight 0.2). The default mode is correct. \\
\textbf{\pkg{crepes}} 0.9.1 & The CRPS of normalised predictive systems multiplies one term by the difficulty estimate $\sigma$ twice. & The reported accuracy score is wrong: a true CRPS of 2.19 is reported as 7.19 at $\sigma=5$. Intervals are not affected. \\ \bottomrule
\end{tabular}
\end{table}

Table~\ref{tab:faults} lists them. Two mechanisms are worth separating here; the fault in \pkg{TorchCP}'s SAPS is explained in the next subsection.

\emph{Too few calibration points, and no warning.} With $n$ calibration errors the largest one covers with probability $n/(n+1)$, so no finite interval reaches 90\% unless $n\ge 9$. The correct answer with fewer errors is an infinite interval, or a refusal. Both \pkg{statsforecast} and \pkg{mlforecast} default to $n=2$ and return a finite interval labelled 90\% that cannot cover more than about $2/3$ of the time. \pkg{Darts} with a calibration length of 3 returns the largest residual, which cannot cover more than $3/4$. These intervals are incorrect, and none of the three packages tells the user: we ran each with Python warnings switched on and none was raised.

\emph{An uncorrected quantile.} For $n\ge 9$ the remaining shortfall in \pkg{statsforecast} and \pkg{mlforecast} comes from taking the ordinary empirical 90\% quantile of the errors (linear interpolation, as in \code{numpy.quantile}) in place of the $\lceil (n+1)(1-\alpha)\rceil$-th smallest. For exchangeable errors this covers about $(1+0.9(n-1))/(n+1)$ (exactly if the errors are uniformly distributed, approximately otherwise): 0.83 at $n=10$, 0.86 at $n=20$, 0.88 at $n=50$, in line with Table~\ref{tab:nixtla}. The \pkg{statsforecast} behaviour had been reported by a user before our audit (issue 1349) and the default was not changed.

\subsection{Five libraries module by module}

Table~\ref{tab:forensic} lists what the module-by-module reading found, in addition to the two items of Table~\ref{tab:faults} that concern \pkg{TorchCP} and \pkg{crepes}. The last column says whether the fault was already on the project's issue tracker when we looked; ``no'' means we found no report among the issue titles on that date, which is not proof that nobody reported it.

{\small
\begin{longtable}{@{}>{\raggedright\arraybackslash}p{6.0cm}>{\raggedright\arraybackslash}p{6.3cm}>{\raggedright\arraybackslash}p{2.0cm}@{}}
\caption{Faults found in the module-by-module audits. Coverage is at a 90\% target and miss rates at a 10\% target unless stated.}\label{tab:forensic}\\ \toprule
Fault & Measured & On tracker \\ \midrule \endfirsthead
\toprule Fault & Measured & On tracker \\ \midrule \endhead
\multicolumn{3}{@{}l}{\textbf{\pkg{MAPIE} 1.6.0}}\\[2pt]
ACI clips its working level to $[0,1]$; this is not the algorithm of \cite{gibbs2021} & Miss rate 13.7\%, 19.8\% and 33.4\% on three drifting sequences, where the theorem allows 10.6\%; unclipped 10.1\%. On i.i.d.\ data 10.1\% clipped and 10.0\% unclipped & no \\
APS and RAPS with \code{include\_last\_label=False} & 81.5\% and 83.6\%; the docstring says coverage may be higher than the target & issue 1021 \\
Default RCPS bound: the test statistic is run on two halves of the data and the two maxima are added & Bound below the true risk in up to 19.7\% of runs where 10\% is allowed; 12.3\% through the public class & pull request 998, closed unmerged \\
Default Holm procedure stops at an undefined p-value, so learn-then-test returns no threshold & No threshold in 161 of 300 runs; Bonferroni on the same data, 0 of 300. Validity not affected & no \\[4pt]
\multicolumn{3}{@{}l}{\textbf{\pkg{crepes} 0.9.1}}\\[2pt]
Out-of-bag calibration with k-nearest-neighbour difficulty estimates counts each training object as its own neighbour & 83.6\% with $k=5$; 88.1\% with the default $k=25$ & no \\
A test object in a Mondrian category absent from calibration gets the interval $[0,0]$ & Every such object, no warning & issue 54 \\
\code{MondrianCategorizer} fitted with \code{learner=} raises \code{NameError} when applied & Every call & no \\
\code{predict(..., seed=)} leaves NumPy's global random state reset to the seed & Every call with a seed & no \\[4pt]
\multicolumn{3}{@{}l}{\textbf{\pkg{PUNCC} 0.9.3}}\\[2pt]
APS and RAPS cap the size of a prediction set at the number of test rows in the call & 45\% with one row per call, 68\% with two, 89\% with five & no \\
Box-wise object detection uses the wrong sign for two sides of the box & True box inside the ``outer'' box in 0\% of cases when the detector's boxes are too small & no \\[4pt]
\multicolumn{3}{@{}l}{\textbf{\pkg{TorchCP} 1.2.1}}\\[2pt]
CQR-R: the correction is scaled by a tensor that is still zero, which leaves a negligible correction & 57.6\%, the same as the raw quantile band to the reported precision; CQR on the same data 90.1\% & no \\
ACI holds its level inside $[1/(n+1),\,0.9999]$ & Miss rate 14.0\% with random-walk errors; bound 10.6\%; unclipped 9.9\% to 10.1\% & no \\
AgACI: no state is carried between steps and all experts always get the same weight & Miss rate 44.5\%, 38.2\% and 27.5\% on three drifting sequences; 10.0\% on i.i.d.\ data & no \\
NorABS fails on any batch of more than one row; used as its docstring says, each calibration row is its own neighbour & \code{RuntimeError}; 83.6\% with $k=5$, computed from the scores because the interval call fails & no \\
\code{ClusteredPredictor} returns every label for classes without a cluster & Valid; mean set size 34.3 against 6.0 for plain split with 30 rare classes in 100 & no \\
\code{WeightedPredictor} sorts the weights separately from the scores & 98.2\% and 99.2\% under covariate shift; sets 41\% to 62\% larger than correct weighting & issue 126 \\
\code{ConformalLM} with a scaler scales calibration scores twice and test scores once & Risk above its limit in 25.5\% of calibration draws where 5\% is allowed (synthetic scores) & no \\
\code{CDLoss} has the sign reversed; the metric \code{SSCL} sums over the wrong axis & Mean set size 3.38 before training, 5.83 after; \code{SSCL} returns 101 for a quantity in $[0,1]$ & no \\[4pt]
\multicolumn{3}{@{}l}{\textbf{\pkg{conformal-tights} 0.5.0}}\\[2pt]
No conformal step below 1{,}414 training rows: the interval is a quantile regression fitted on the calibration set & 80.6\% at a 90\% target and 84.8\% at 95\% with 300 rows, mean of eight public datasets; 84.2\% and 89.9\% with 1{,}000 rows (seven datasets) & not checked \\
With enough rows, the narrower of two calibrated intervals is returned for each test point & 87.2\% and 93.3\% with 5{,}000 rows, mean of five datasets; at the 90\% target each of the two intervals alone gives 90.6\% and 91.3\% & not checked \\
The second calibration stage ignores sample weights & With one fifth of the rows corrupted and given weight zero, the 90\% interval is three times wider than a fit on the clean rows alone and covers 99.7\% & not checked \\
\pkg{Darts} forecaster: calibration rows are drawn at random from the lagged series & One step ahead on the project's own example (2{,}184 test hours): 86.8\% at 90\%, 92.0\% at 95\% & not checked \\
\code{predict\_interval} edits a stored array, which changes what a later \code{predict\_quantiles} call at the same levels returns & 13 of 20 fits & not checked \\ \bottomrule
\end{longtable}}

Five patterns recur, and they are worth more than the list.

\emph{A clipped level.} In adaptive conformal inference the working level moves by a fixed step after every outcome and is allowed to leave $[0,1]$; at or below zero the interval is the whole line. The guarantee of \cite{gibbs2021}, a long-run miss rate within $(\max(\alpha_1,1-\alpha_1)+\gamma)/(\gamma T)$ of the target for any sequence, needs every miss to lower the level by the full step. \pkg{MAPIE} and \pkg{TorchCP} both clip the level. A miss near the boundary is then charged less than a full step, and on a drifting series the next interval can miss again. On stationary and mildly dependent data the clip rarely binds (on i.i.d.\ data the clipped loop missed 10.1\% and the unclipped one 10.0\%), which is why ordinary tests do not show it. Validity under any drift, the property that distinguishes the method, is the part that is lost. Our implementation does not clip, and returns infinite intervals when the algorithm calls for them.

\emph{An object that is its own neighbour.} Normalised conformal regression divides residuals by a difficulty estimate, often built from nearest neighbours. If the neighbours of a calibration object are searched in a set that contains the object, its difficulty is computed from its own residual and its normalised score is too small. The \pkg{crepes} package does this in the out-of-bag recipe of its own notebook, and \pkg{TorchCP} does it when NorABS is used as its docstring describes. Both give 83.6\% at a 90\% target with five neighbours.

\emph{Calibration and prediction that disagree.} \pkg{TorchCP}'s non-randomised SAPS (below), \pkg{MAPIE}'s APS and RAPS with \code{include\_last\_label=False}, and \pkg{TorchCP}'s language-model predictor with a scaler all score calibration data one way and test data another. The rank argument needs the same score on both sides.

\emph{Results that depend on how \code{predict} is called.} \pkg{PUNCC}'s APS and RAPS return different sets for the same row depending on how many other rows are in the call. In \pkg{conformal-tights} the quantiles returned differ depending on whether an interval was requested first.

\emph{A conformal step that is not there.} \pkg{TorchCP}'s CQR-R and \pkg{conformal-tights} on fewer than 1{,}414 rows return what is in effect an uncorrected model output under a conformal name. For \pkg{conformal-tights} the project page lists ``requires only a small number of calibration examples'' among its features; the package's own coverage test passes at 97\% of the requested level.

\emph{Two scores for one class.} \pkg{TorchCP}'s non-randomised SAPS gives the top-ranked class a calibration score of 0 and a prediction score of $p_{\max}$; all other ranks agree. When the true label is the top class on more than a fraction $1-\alpha$ of calibration points, the threshold is 0 or close to it, and at prediction time even the top class is excluded whenever $p_{\max}$ exceeds it. When top-1 accuracy is well below $1-\alpha$ the threshold is set by the higher ranks, where the two scores agree, and the fault is invisible. That is why Table~\ref{tab:sets} shows 34\% coverage with 10 classes and correct coverage with 100 and 1{,}000, where the mean set holds 9 and 81 classes. The figures in Table~\ref{tab:faults} come from a separate run with 10 classes in which accuracy was varied through the scale of the logits (Appendix~\ref{app:details}).

A later, wider test of the same option found that once top-1 accuracy passes the target level the sets are empty: coverage 0.0\% at accuracies of 0.93 and 0.98.

\emph{The second calibration stage of \pkg{conformal-tights}.} The package fits a non-crossing linear quantile regression on 70\% of its calibration rows and then shifts each quantile by a constant estimated on the other 30\%. Three things weaken that shift. It is applied only if those 30\% hold at least 128 rows, which needs 1{,}414 training rows; between 425 and 1{,}413 rows the 30\% are set aside and then not used at all. It is an ordinary empirical quantile, without the finite-sample correction, and it ignores sample weights. And for levels between the lowest and the highest it is clipped, to keep the lines from crossing on the calibration points: on 16 fits with 5{,}000 rows the data asked for shifts of 0.035 to 0.052 residual standard deviations at the levels 0.25 to 0.75, and shifts of 0.005 to 0.019 were applied. The interval finally returned is, for each test point, the narrower of one built from absolute residuals and one built from relative residuals. Each of the two is close to its target when there are enough rows; choosing the narrower point by point is not. To be fair to the method: with 5{,}000 rows its intervals had a better interval score than split conformal and CQR built on the same data on four of five datasets. It trades one to five points of coverage for sharpness, and does not say so.

Smaller items. In \pkg{crepes} the code takes the $\lfloor (n+1)(1-c)\rfloor$-th largest residual for confidence $c$; at $c=0.9$ the product falls just below a whole number when $n+1$ is a multiple of ten (Section 3.1), so it uses one rank too high in 40 of the 1{,}240 cases of Table~\ref{tab:exact} and returns an infinite interval for $n=9$. \pkg{PUNCC} raises a \code{TypeError} with a single calibration point (the ten errors in Table~\ref{tab:exact}; at two of them a finite interval is valid) and its EnbPI raises one when residuals are updated after every observation. Nixtla's \pkg{conformalforecast} refuses a level when $n$ is exactly the smallest size at which it is feasible, for example 9 points at 90\% (the ten refusals in Table~\ref{tab:exact}). \pkg{MAPIE}'s RAPS failed with 1{,}000 classes and 5{,}000 calibration points because it splits the calibration data by class and needs two points in every class; this is an error message, not a coverage fault. R's \code{scp} takes one rank too high at a few calibration sizes, and the default integrator scale in R's \code{acmcp} is the largest error of the whole history, which includes errors not yet observed.

Further small items. \pkg{MAPIE}: \code{SplitConformalRegressor} raises a consistency error on float32 inputs with targets centred near zero (200 of 200 random datasets; reported in 2023 and 2024 and closed with a workaround); \code{allow\_infinite\_bounds=True} is ignored with the default symmetric score; \code{spiegelhalter\_p\_value} is one-sided. \pkg{crepes}: the rank is computed in floating point and is one too high at about a tenth of calibration sizes at a 90\% level and a fifth at 80\%, which widens intervals and never narrows them; \code{WrapRegressor.fit} returns \code{None}. \pkg{TorchCP}: the ACI miss indicator counts a value exactly on the bound as a miss; \code{ACIPredictor.train} calibrates on the training data. \pkg{conformal-tights}: quantiles cross on 0.2\% of new points with five levels, because the constraint is imposed on calibration points only, and up to 1\% of returned intervals have their lower end above their upper end on small samples; the first quantile prediction with five levels on 5{,}000 rows takes about 50 seconds, because the non-crossing fit is a linear programme solved at prediction time.

\subsection{What was correct}

The basic routines of the four main libraries are sound. \pkg{MAPIE}'s split, cross and jackknife regression, CQR and LAC met their target in 78 of 84 configurations and refused to run in the other six because the calibration set was too small, which is safe; its two shared quantile functions returned the exact conformal rank in 37{,}581 of 37{,}581 cases, and its public interval call refuses a level that the calibration set cannot support. Its conformal risk control, its RCPS with Hoeffding and Bernstein bounds, its conditional conformal regressor, its exchangeability tests and its calibrators all behaved as documented. The \pkg{crepes} package met its level in all 114 coverage settings for standard, normalised and Mondrian regressors, predictive systems and classifiers; its p-values are uniform and its online mode agrees with a batch refit. On \pkg{PUNCC}'s regression side every method met its level and the threshold was the exact rank at all 9{,}835 sizes tested; LAC, class-wise LAC and anomaly detection were correct. \pkg{TorchCP}'s rank was exact in 9{,}835 of 9{,}835 cases and its classification scores with the split predictor covered 89.7\% to 89.9\% at 90\%, with no dependence on batch size; its class-conditional predictor, p-values, graph scores and predictive distribution were correct. The packages \pkg{venn-abers} 1.5.4 and \pkg{online-cp} 0.4.0 were correct on everything tested.

\subsection{Valid but larger than needed}

\pkg{MAPIE}'s default APS and RAPS are not randomised and covered 97.5\% and 96.9\% at a 90\% target with 10 classes, with sets about 20\% larger than the randomised versions. \pkg{PUNCC}'s non-randomised APS covered 99.2\% in our coverage suite (it is not part of Table~\ref{tab:sets}). These are choices, not errors, but the defaults cost efficiency. Three items of Table~\ref{tab:forensic} belong here too: \pkg{TorchCP}'s \code{ClusteredPredictor} and \code{WeightedPredictor} keep coverage and return sets far larger than the methods they are named after, and \pkg{MAPIE}'s default Holm procedure keeps validity and often returns no threshold for a target the model can reach.

\subsection{Limits of the audit}

Coverage tests use synthetic data, except that \pkg{conformal-tights} was also run on eight public datasets. Parts not covered are listed in each report; the main ones are \pkg{MAPIE}'s semantic-segmentation controller and experimental module, \pkg{PUNCC}'s deep-learning backends, \pkg{TorchCP}'s language-model module with a real language model, and \pkg{conformal-tights}' issue tracker. All findings are for the versions named, on the dates named; later releases may differ. The fault in \pkg{MAPIE}'s RCPS bound was first described by an outside contributor in a pull request to that project; we found it independently from the code and learned of the pull request afterwards.

\section{Benchmarks}\label{sec:bench}

Timings are from a 2-core Linux machine with Numba installed, single runs or best of three, on 20--21 November 2026. Appendix~\ref{app:details} lists the environment and the controls applied, and Appendix~\ref{app:extra} reports further experiments: latency one row at a time with a model of controlled cost, stress cases for Venn--Abers, the approximate forest mode, very large label sets, and an ablation of CIR.

\begin{table}[t]\centering\small
\caption{Speed against other packages and against code released with the papers. ``Same'' means the outputs were compared and found equal.}\label{tab:speed}
\setlength{\tabcolsep}{4pt}\footnotesize
\begin{tabular}{@{}p{3.4cm}p{3.3cm}rrrl@{}}\toprule
Method & Against & Theirs & Ours & Factor & Same \\ \midrule
Venn--Abers, $10^6/10^6$ & \pkg{venn-abers}, \pkg{MAPIE} & 75.0 s & 0.366 s & 205 & yes \\
Venn--Abers, $10^5/10^6$ & \pkg{venn-abers} & 2.32 s & 0.079 s & 29 & yes \\
CPS intervals, $10^5/10^6$ & \pkg{crepes} & 79.2 s & 0.0098 s & 8{,}100 & yes \\
CPS CRPS, 5{,}000/5{,}000 & \pkg{crepes} & 0.309 s & 0.00047 s & 656 & yes$^{a}$ \\
Split regression, $2{\times}10^5/10^6$ & \pkg{PUNCC}, \pkg{crepes}, \pkg{MAPIE} & 4.2--8.9 ms & 1.9 ms & 2--5 & yes \\
CQR, $10^5/10^6$ & \pkg{MAPIE}; \pkg{PUNCC} & 19.7; 10.5 ms & 3.2 ms & 6; 3 & yes \\
CTI & authors' code & 431 ms & 23 ms & 19 & yes \\
CIR; CIR+ & authors' code & 1.85; 2.93 s & 0.12; 0.14 s & 16; 22 & yes; no \\
CV+ prediction & \pkg{MAPIE} & 1.95--4.99 s & 0.12--0.18 s & 17--27 & coverage \\
APS, RAPS, SAPS (Table~\ref{tab:sets}) & \pkg{TorchCP}, \pkg{MAPIE}, \pkg{PUNCC} & 0.016--1.9 s & 0.008--0.36 s & 2--7 & see$^{c}$ \\
Quantile forest, fit$^{b}$ & \pkg{quantile-forest} & 11.5--215 s & 1.9--27 s & 6--9 & loss \\
MSTL, $10^5$ points (Table~\ref{tab:mstl}) & \pkg{statsforecast}; R \code{forecast} & 19.3 s; --- & 0.098 s & 197$^{d}$; 5 & yes \\
ACI, replay & \pkg{MAPIE} & 0.41 ms/step & 0.2 $\mu$s/step & $\sim$2{,}000 & coverage \\
ACI, one row at a time & \pkg{MAPIE} & 0.57 ms/step & 0.09 ms/step & 6--8 & coverage \\
ACI & authors' code & 114 $\mu$s/step & 0.40 $\mu$s/step & 286 & n/c \\
Conformal PID & authors' code & 10--14 $\mu$s/step & 0.09--0.13 $\mu$s/step & 106--112 & yes \\
DtACI, update step & authors' R code & 41 $\mu$s/step & 0.62 $\mu$s/step & 66 & n/a \\
EnbPI, replay & \pkg{MAPIE} & 5{,}454 $\mu$s/step & 11.6 $\mu$s/step & 400--470$^{e}$ & coverage \\
EnbPI, one row at a time & \pkg{MAPIE} & 5{,}800 $\mu$s/step & 2{,}200 $\mu$s/step & 2.5 & coverage \\
SPCI & paper's loop & 29 ms/step & 3.3 ms/step & 9 & n/c \\
Multi-step split & R \code{scp} & 42 s & 0.010 s & $>$4{,}000 & yes \\
AcMCP, no scorecaster, 1{,}000$\times$6 & R \code{acmcp} & 3.9 s & 0.8 ms & $\sim$5{,}000 & yes \\
AcMCP, with scorecaster, 1{,}000$\times$6 & R \code{acmcp} & 107 s & 2.5 s & 43 & close \\ \bottomrule
\end{tabular}

\smallskip{\footnotesize $^{a}$ Without difficulty estimates; with them \pkg{crepes}' value is wrong (Table~\ref{tab:faults}). $^{b}$ Measured on a 16-core server, 100{,}000 and 1{,}000{,}000 training rows; ``loss'' means equal pinball loss. $^{c}$ LAC sets are identical across packages; randomised scores use each package's own random draws, so coverage and set size are compared, not the sets. $^{d}$ At \pkg{statsforecast}'s default settings; 6--7 at equal settings. $^{e}$ Three runs gave 399, 402 and 470. Sizes $a/b$ are calibration and test points; $a\times b$ is time steps by horizons.}
\end{table}

Table~\ref{tab:speed} gives the main results; Appendix~\ref{app:tables} has the full tables. Six qualifications matter.

\paragraph{Replay and live use are different measurements.} For ACI and EnbPI the large factors against \pkg{MAPIE} are for replaying a history whose forecasts or features are all available, which is what a backtest does: \pkg{pyconformal} calls the model once for all rows. \pkg{MAPIE}'s interface calls the model at every step. When both are fed one row at a time, the cost of a single-row model call dominates and the factors are 6--8 for ACI and about 2.5 for EnbPI. Both rows are in the table and neither should be quoted without its label. To separate the conformal step from the model, Table~\ref{tab:live} repeats the one-row-at-a-time measurement with a stub model of zero and of fixed cost: with a zero-cost model the median step takes 620~$\mu$s in \pkg{MAPIE} and 9~$\mu$s in \pkg{pyconformal}.

\paragraph{Venn--Abers gains depend on the scores.} In the run of Table~\ref{tab:va} the factor is 205 at a million calibration points with continuous scores, 11--29 at $10^4$--$10^5$, 6--17 on uninformative or anti-correlated scores, and 3--6 when scores are rounded to two decimals. Without Numba it is 25 at a million points. Timings of the reference at this size varied between runs (61.5 and 75.0 seconds); elsewhere we quote the lower factor from the faster run, 188.

\paragraph{Memory.} Table~\ref{tab:mem} gives the peak extra memory of one call for the four methods we measured. The large reductions are where another package builds a matrix that is not needed: CV+ prediction (3 megabytes against 1.3--3.1 gigabytes) and the CRPS (0.4 against 383 megabytes). Intervals from predictive systems need about six times less. Venn--Abers and the distribution function need about what the reference needs. Memory of the other methods was not measured.

\paragraph{Small margins are small.} Split conformal regression takes milliseconds in every package at a million test points.

\paragraph{Comparisons with R} set compiled Python against interpreted R. The AcMCP rows of Table~\ref{tab:speed} are for 1{,}000 steps and 6 horizons, the largest size at which we ran R. At 3{,}000 steps and 12 horizons, the size used for multi-step split conformal prediction, \pkg{pyconformal}'s AcMCP with the scorecaster takes 127 seconds; R was not run at that size. The scorecaster is the slowest part of the package.

\paragraph{Warm-up and the ``Same'' column.} Compiled kernels are called once before timing, so compilation is excluded from every figure; it takes from a fraction of a second to a few seconds per kernel on first use and is cached on disk afterwards. In Table~\ref{tab:speed}, ``yes'' means outputs were compared and equal, ``coverage'' that only coverage was compared, ``n/c'' not compared and ``n/a'' not comparable.

\subsection{Statistical efficiency}

On a skewed outcome with 100 predicted quantile levels, 20{,}000 calibration and 200{,}000 test points and a 90\% target, mean interval length was 5.59 for CQR, 5.18 for CIR, 5.02 for CIR+ and 4.95 for CTI (a set that may be a union of intervals), with coverage 0.899--0.909. Table~\ref{tab:cirabl} repeats the comparison with standard errors on three outcome distributions and two levels. Randomised APS in \pkg{pyconformal} gave sets of mean size 1.73 at 89.9\% coverage where \pkg{MAPIE}'s default APS gave 2.05 at 97.5\%.

\subsection{Real series}\label{sec:real}

\begin{figure}[ht]
\centering
\includegraphics[width=0.85\linewidth]{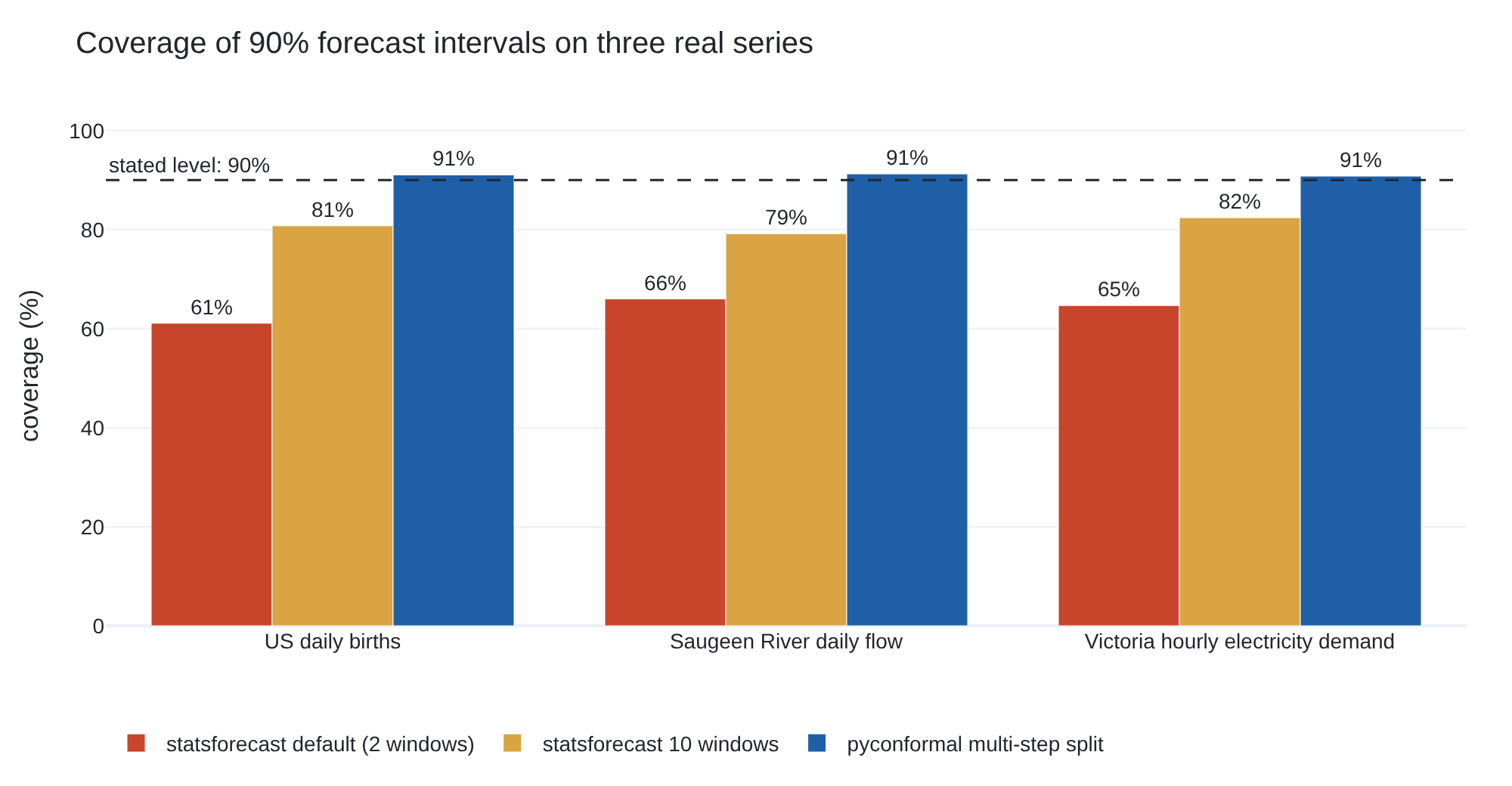}
\caption{Coverage of 90\% forecast intervals on three real series. The point forecasts are the same for every bar.}\label{fig:real}
\end{figure}

\begin{figure}[ht]
\centering
\includegraphics[width=0.92\linewidth]{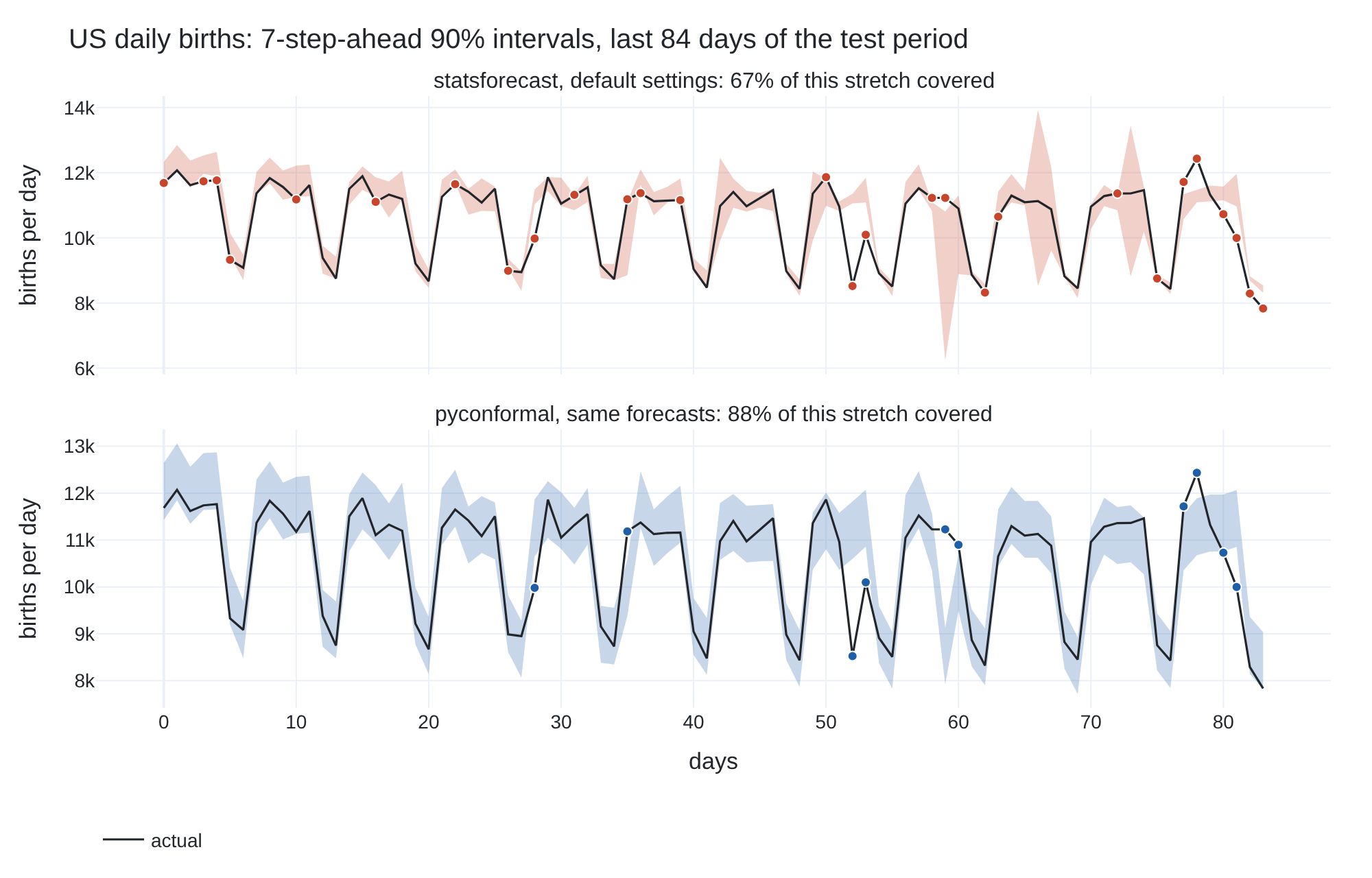}
\caption{US daily births, forecasts 7 days ahead, 90\% intervals over the last 84 days of the test period. Top: \pkg{statsforecast} with its default settings. Bottom: \pkg{pyconformal} on the same point forecasts. Dots mark outcomes outside the interval.}\label{fig:births}
\end{figure}

\begin{figure}[ht]
\centering
\includegraphics[width=0.92\linewidth]{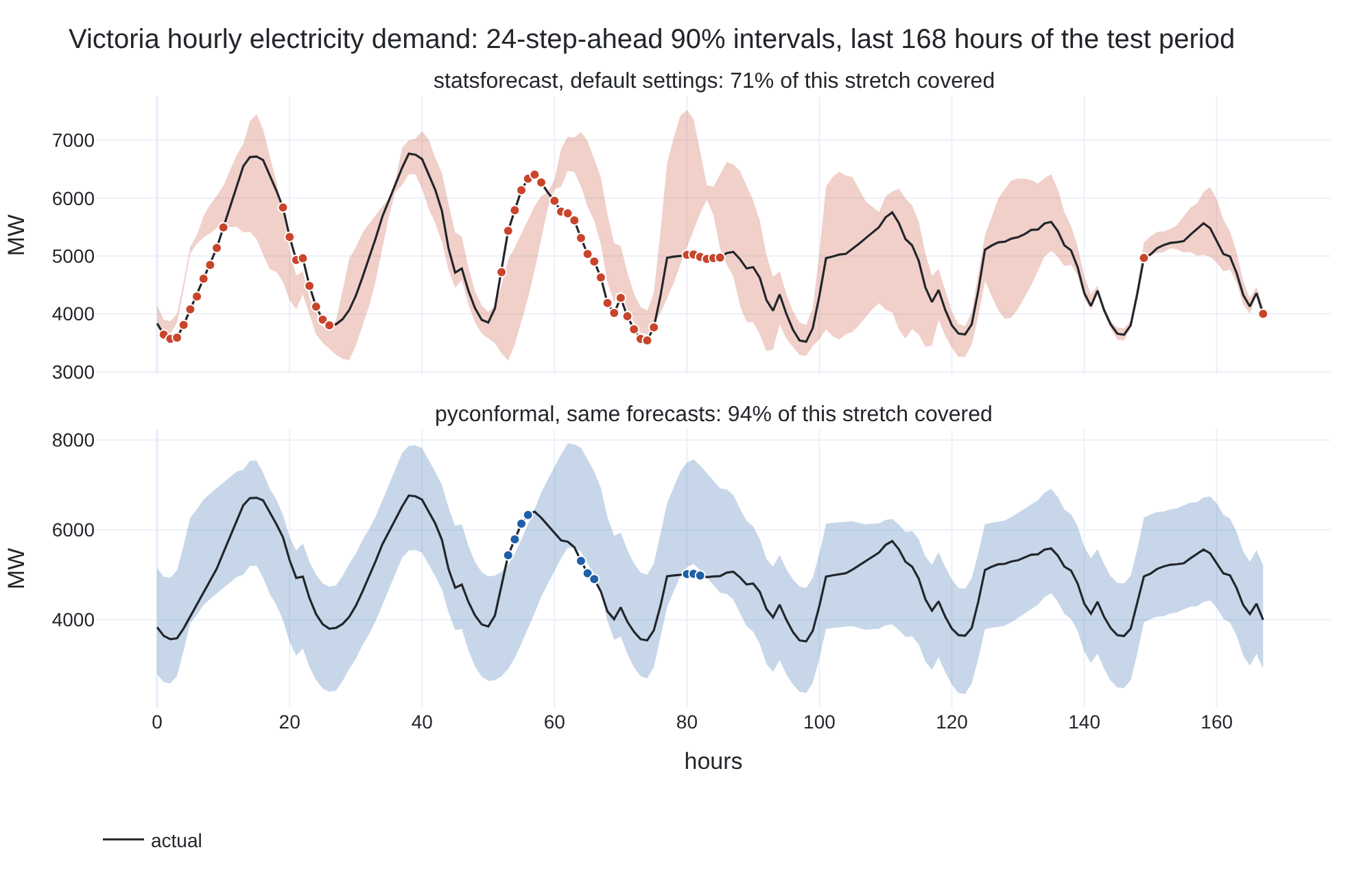}
\caption{Victoria electricity demand, forecasts 24 hours ahead, 90\% intervals. Top: \pkg{statsforecast} with its default settings. Bottom: \pkg{pyconformal} on the same point forecasts. Dots mark outcomes outside the interval.}\label{fig:power}
\end{figure}

We took three public series from the Monash archive \cite{godahewa2021}: daily births in the United States (7{,}305 days), the daily flow of the Saugeen River (23{,}741 days) and hourly electricity demand in Victoria, Australia. At each of 1{,}500 consecutive forecast origins \pkg{statsforecast} 2.1.1 was fitted to the most recent history (364 days, or 28 days of hourly data) and forecast 7 days or 24 hours ahead, with a seasonal naive model for births and electricity and a naive model for the river. We recorded its own conformal intervals at a 90\% level, with the default of 2 calibration windows and with 10, and gave the same point forecasts to \code{MultiStepSplitConformal}, which for each target uses only the forecast errors already known when the forecast was made. Coverage is counted over the same targets for every method, after the first 100 origins (Table~\ref{tab:real}, Figure~\ref{fig:real}). Figures~\ref{fig:births} and~\ref{fig:power} show a stretch of two of the series: with two calibration errors the default intervals are narrow in most places and miss often, and now and then one large error makes them very wide.

\begin{table}[ht]
\centering\small
\caption{Coverage of 90\% forecast intervals on three real series, all horizons pooled, about 1{,}390 targets per horizon. Width is the median interval length in the units of the series. Bold marks the best result for each series: the narrowest intervals among those that cover within one point of the 90\% target.}\label{tab:real}
\begin{tabular}{llrr}
\toprule
Series & Intervals & Coverage & Median width \\
\midrule
US births & \pkg{statsforecast}, default & 61.1\% & 518 \\
 & \pkg{statsforecast}, 10 windows & 80.8\% & 890 \\
 & \pkg{pyconformal}, multi-step split & \textbf{91.0\%} & \textbf{1{,}234} \\
 & \pkg{pyconformal}, AcMCP, no scorecaster & 89.9\% & 2{,}025 \\
Saugeen River & \pkg{statsforecast}, default & 66.0\% & 10.7 \\
 & \pkg{statsforecast}, 10 windows & 79.2\% & 29.0 \\
 & \pkg{pyconformal}, multi-step split & \textbf{91.3\%} & \textbf{74.8} \\
 & \pkg{pyconformal}, AcMCP, no scorecaster & 89.7\% & 233 \\
Victoria electricity & \pkg{statsforecast}, default & 64.7\% & 1{,}015 \\
 & \pkg{statsforecast}, 10 windows & 82.4\% & 1{,}839 \\
 & \pkg{pyconformal}, multi-step split & \textbf{90.8\%} & \textbf{2{,}393} \\
 & \pkg{pyconformal}, AcMCP, no scorecaster & 89.3\% & 5{,}941 \\
\bottomrule
\end{tabular}
\end{table}

The default covers 61\% to 66\%, which matches the synthetic result in Table~\ref{tab:faults}: two errors per horizon cannot give a valid finite 90\% interval. With 10 windows coverage is about 80\%, because the interval comes from an interpolated quantile of the errors, which lands just above the second largest of ten, where the largest is needed. Multi-step split conformal reaches the stated level with intervals that are wider, as a 90\% interval must be. It is plain split conformal prediction applied per horizon and recalibrated at every origin on all errors known so far, which takes 30 to 100 milliseconds for a whole series because no model is refitted; no adaptive or weighted method was needed to reach the level. The \pkg{statsforecast} library uses the same idea, and the gap comes from the number of errors used and the quantile taken. AcMCP without its scorecaster also reaches the level, but its intervals are 1.6 to 3 times wider than multi-step split on these series. With \pkg{statsforecast}'s AutoETS in place of the naive model on the river series (1{,}093 targets per horizon), coverage was 63.4\%, 79.5\% and 90.0\% for the default, 10 windows and \pkg{pyconformal}. Forecast errors are not exchangeable, so these are observed frequencies with no guarantee behind them; with only the last 200 errors per horizon, multi-step split covered 88.4\% to 89.6\%.

\FloatBarrier
\section{Limitations}\label{sec:limits}

\begin{itemize}
\item All coverage experiments except Section~\ref{sec:real} and the real-data runs of the \pkg{conformal-tights} audit use synthetic data. Time-series methods have no finite-sample guarantee under dependence, and we observed 85\% coverage for SPCI and 89\% for EnbPI at a 90\% target on strongly autocorrelated errors.
\item Peak memory was measured for CV+, predictive systems and Venn--Abers only (Table~\ref{tab:mem}), in one run each. Elsewhere we make no claim about memory.
\item Timings come from one 2-core machine. The quantile forest alone was also measured on a 16-core server.
\item CIR differs from its authors' code in two ways, on purpose. Their code scores calibration points with a window that grows from the mode and predicts with the shortest window of the resulting number of intervals; these are not always the same window, so the rank argument does not apply. Our code uses the growing window on both sides by default and offers the authors' rule as \code{window="shortest"}, which returned intervals identical to theirs on all 200{,}000 test points of our benchmark. The difference matters: on a bimodal outcome the authors' rule covered 88.3\% at a 90\% target and 76.7\% at an 80\% target in their own code, while the default covered 90.5\% and 80.4\% (Table~\ref{tab:cirabl}). For CIR+ their code scales the added length by 100 times the largest calibration outcome; ours scales by the sample's predicted range.
\item AcMCP with the scorecaster agrees with R to a median of $10^{-5}$, not exactly, because both packages fit the MA model with an iterative optimiser. Where we examined larger differences our fit had the higher likelihood.
\item Not yet included: conformal risk control, localised and clustered conditional methods, conformal anomaly detection, exchangeability testing.
\item The Venn--Abers tangent walk has no proved worst-case bound. On nine stress configurations, including alternating labels, blocks of equal labels, heavy ties and geometrically spaced scores, time grew by a factor of 11 to 18 for ten times the data, and the output equalled the reference implementation exactly (Table~\ref{tab:vastress}).
\item The real-data study in Section~\ref{sec:real} covers three series and simple forecasters; it is an illustration, not a benchmark. Timings were not repeated on a second machine.
\item Weighted conformal predictive systems are new in version 0.1.6; they were tested against their definition, by simulation under a known covariate shift, and against \pkg{crepes-weighted} 0.1.3, the code released with \cite{jonkers2024}. The distribution functions of the two agree to $10^{-15}$. The quantile rules differ: with equal weights ours returns the quantiles of the unweighted system, while \pkg{crepes-weighted} takes ranks one to two places further out (ranks 3 and 96 where the unweighted system takes 5 and 95, with 99 calibration points) and returns infinite intervals more often; both rules kept their level in simulation. Weighted ranks use floating point with a tolerance, not exact arithmetic. The exact rank routine, \code{conformal\_quantile}, is a public function, so other code can reuse it without the rest of the package.
\end{itemize}

\section{Availability}

The package \pkg{pyconformal} is released under the MIT licence at \url{https://github.com/valeman/pyconformal} and on PyPI as \code{pyconformal}; \code{pip install "pyconformal[fast]"} adds Numba. It requires Python 3.10 or later and NumPy 1.22 or later. The repository contains the test suite, continuous integration, and the benchmark scripts behind the timings in this paper, including short self-contained scripts that reproduce the faults in Tables~\ref{tab:faults} and~\ref{tab:forensic} using only the package concerned.

\clearpage
\bibliographystyle{plainnat}
\bibliography{references}
\clearpage

\appendix
\section{Full result tables}\label{app:tables}

All runs used the releases current on 20 November 2026. Each table names its script in the repository's \code{benchmarks} folder.

\begin{table}[h]\centering\small
\caption{The split-conformal threshold over 1{,}240 cases: $n = 1,\dots,120, 199, 200, 999, 1000$ crossed with $\alpha \in \{0.01, 0.02, 0.05, 0.1, 0.15, 0.2, 0.25, 0.3, 0.5, 0.9\}$ (\code{exactness.py}). In 184 of the cases no finite interval is valid; a package may then return an infinite interval or refuse, and both are safe. Each package is called through its public interval method, except \pkg{TorchCP}, whose threshold function is called directly. The entries are explained in Section~\ref{sec:audit}.}\label{tab:exact}
\resizebox{\textwidth}{!}{\begin{tabular}{@{}lrrrrrrr@{}}\toprule
 & \multicolumn{3}{c}{Correct or safe} & \multicolumn{4}{c}{Not correct} \\ \cmidrule(lr){2-4}\cmidrule(l){5-8}
Package & Exact, finite & Infinite & Refused, none valid & Too narrow & Too wide & Refused, one valid & Error \\ \midrule
\pkg{pyconformal} & 1{,}056 & 184 & 0 & 0 & 0 & 0 & 0 \\
\pkg{TorchCP} 1.2.1 & 1{,}056 & 184 & 0 & 0 & 0 & 0 & 0 \\
\pkg{MAPIE} 1.6.0 & 1{,}056 & 0 & 184 & 0 & 0 & 0 & 0 \\
\pkg{PUNCC} 0.9.3 & 1{,}054 & 0 & 176 & 0 & 0 & 0 & 10 \\
\pkg{conformalforecast} 0.1.0 (Nixtla) & 1{,}046 & 0 & 184 & 0 & 0 & 10 & 0 \\
\pkg{crepes} 0.9.1 & 1{,}016 & 184 & 0 & 0 & 40 & 0 & 0 \\ \bottomrule
\end{tabular}}
\end{table}

\begin{table}[h]\centering\small
\caption{Coverage of \pkg{statsforecast} 2.1.1 and \pkg{mlforecast} 1.1.0 conformal intervals at a nominal 90\% on 4{,}000 independent series (\code{nixtla\_coverage.py}), and of a plain empirical quantile against the exact rule in simulation (\code{exact\_rule\_simulation.py}). ``Distribution'' and ``error'' are the two options of these packages, \code{conformal\_distribution} (signed errors) and \code{conformal\_error} (absolute errors). The plain quantile uses linear interpolation.}\label{tab:nixtla}
\begin{tabular}{@{}lrrrr@{}}\toprule
 & \multicolumn{2}{c}{\pkg{statsforecast}} & \multicolumn{2}{c}{\pkg{mlforecast}} \\
Calibration windows & distribution & error & distribution & error \\ \midrule
2 (default) & 0.634 & 0.645 & 0.634 & 0.645 \\
5 & 0.776 & 0.781 & 0.773 & 0.781 \\
10 & 0.823 & 0.830 & 0.824 & 0.832 \\
20 & 0.860 & 0.862 & 0.862 & 0.865 \\ \bottomrule
\end{tabular}

\medskip
\begin{tabular}{@{}lrl@{}}\toprule
Calibration errors & Plain quantile & Exact rule \\ \midrule
2 & 0.648 & 1.000 (infinite interval) \\
5 & 0.787 & 1.000 (infinite interval) \\
9 & 0.830 & 0.900 \\
10 & 0.833 & 0.910 \\
20 & 0.862 & 0.903 \\
50 & 0.885 & 0.902 \\ \bottomrule
\end{tabular}
\end{table}

\begin{table}[h]\centering\small
\caption{Venn--Abers, fit plus predict, in seconds (\code{speed\_venn\_abers.py}). Outputs are identical in every row. One run on 21 November 2026; the last column is the reference time divided by ours with Numba.}\label{tab:va}
\resizebox{\textwidth}{!}{\begin{tabular}{@{}lrrrrrrr@{}}\toprule
Scores & Calibration & Test & \pkg{venn-abers} & \pkg{MAPIE} & Ours & Ours, no Numba & Factor \\ \midrule
Realistic & $10^4$ & $10^5$ & 0.087 & 0.086 & 0.0081 & 0.028 & 11 \\
Realistic & $10^5$ & $10^6$ & 2.317 & 2.258 & 0.0793 & 0.251 & 29 \\
Realistic & $10^6$ & $10^6$ & 74.96 & 75.65 & 0.3662 & 2.985 & 205 \\
Rounded (2 decimals) & $10^4$ & $10^5$ & 0.015 & 0.015 & 0.0033 & 0.004 & 5 \\
Rounded (2 decimals) & $10^5$ & $10^6$ & 0.144 & 0.126 & 0.0254 & 0.027 & 6 \\
Rounded (2 decimals) & $10^6$ & $10^6$ & 0.183 & 0.223 & 0.0687 & 0.100 & 3 \\
Pure noise & $10^4$ & $10^5$ & 0.049 & 0.049 & 0.0078 & 0.024 & 6 \\
Pure noise & $10^5$ & $10^6$ & 0.631 & 0.630 & 0.0902 & 0.286 & 7 \\
Pure noise & $10^6$ & $10^6$ & 5.173 & 4.210 & 0.4717 & 3.865 & 11 \\
Anti-correlated & $10^4$ & $10^5$ & 0.054 & 0.051 & 0.0063 & 0.017 & 9 \\
Anti-correlated & $10^5$ & $10^6$ & 0.592 & 0.618 & 0.0904 & 0.257 & 7 \\
Anti-correlated & $10^6$ & $10^6$ & 4.856 & 5.395 & 0.2939 & 2.157 & 17 \\ \bottomrule
\end{tabular}}
\end{table}

\begin{table}[h]\centering\small
\caption{Conformal predictive systems against \pkg{crepes} 0.9.1, in seconds (\code{speed\_predictive.py}).}\label{tab:cps}
\begin{tabular}{@{}lrrrrr@{}}\toprule
Task & Calibration & Test & \pkg{crepes} & Ours & Factor \\ \midrule
90\% intervals & $10^3$ & $10^4$ & 0.0358 & 0.00015 & 241 \\
90\% intervals & $10^4$ & $10^5$ & 0.620 & 0.0010 & 602 \\
90\% intervals & $10^5$ & $10^6$ & 79.2 & 0.0098 & 8{,}126 \\
Distribution function & $10^3$ & $10^4$ & 0.081 & 0.0019 & 42 \\
Distribution function & $10^4$ & $10^5$ & 1.27 & 0.020 & 64 \\
Distribution function & $10^5$ & $10^6$ & 145 & 0.261 & 556 \\
CRPS & 1{,}000 & 2{,}000 & 0.037 & 0.00016 & 225 \\
CRPS & 5{,}000 & 5{,}000 & 0.309 & 0.00047 & 656 \\ \bottomrule
\end{tabular}
\end{table}

\begin{table}[h]\centering\small
\caption{Prediction sets at a 90\% target: coverage / mean set size / seconds, for 5{,}000 calibration points and 50{,}000 test points (20{,}000 for 1{,}000 classes) (\code{speed\_classification.py}). RAPS uses penalty 0.01 and $k_\text{reg}=5$, SAPS weight 0.2. \pkg{MAPIE}'s RAPS needs two calibration points in every class and raised an error with 1{,}000 classes. Single runs; coverage carries sampling noise of about $\pm 0.002$.}\label{tab:sets}
\begin{tabular}{@{}llll@{}}\toprule
Method, package & 10 classes & 100 classes & 1{,}000 classes \\ \midrule
LAC, ours & 0.909 / 1.23 / 0.001 & 0.905 / 7.58 / 0.011 & 0.897 / 78.7 / 0.062 \\
LAC, \pkg{TorchCP} & 0.909 / 1.23 / 0.007 & 0.905 / 7.58 / 0.018 & 0.897 / 78.7 / 0.146 \\
LAC, \pkg{MAPIE} & 0.909 / 1.23 / 0.010 & 0.905 / 7.58 / 0.064 & 0.897 / 78.7 / 0.293 \\
LAC, \pkg{PUNCC} & 0.909 / 1.23 / 0.133 & 0.905 / 7.58 / 0.321 & 0.897 / 78.7 / 0.895 \\
APS randomised, ours & 0.899 / 1.73 / 0.008 & 0.901 / 8.38 / 0.066 & 0.899 / 78.4 / 0.364 \\
APS randomised, \pkg{TorchCP} & 0.907 / 1.76 / 0.017 & 0.900 / 8.41 / 0.276 & 0.899 / 78.7 / 1.696 \\
APS randomised, \pkg{PUNCC} & 0.901 / 1.74 / 0.129 & 0.899 / 8.37 / 0.293 & 0.899 / 78.4 / 1.139 \\
APS default, \pkg{MAPIE} & 0.975 / 2.05 / 0.043 & 0.923 / 9.07 / 0.309 & 0.900 / 78.9 / 1.832 \\
RAPS randomised, ours & 0.899 / 1.73 / 0.010 & 0.902 / 8.18 / 0.061 & 0.896 / 79.5 / 0.267 \\
RAPS randomised, \pkg{TorchCP} & 0.907 / 1.76 / 0.018 & 0.902 / 8.21 / 0.225 & 0.896 / 79.7 / 1.661 \\
RAPS default, \pkg{MAPIE} & 0.969 / 2.11 / 0.047 & 0.916 / 10.26 / 0.386 & error \\
SAPS randomised, ours & 0.902 / 1.65 / 0.008 & 0.904 / 8.60 / 0.040 & 0.896 / 81.1 / 0.278 \\
SAPS randomised, \pkg{TorchCP} & 0.903 / 1.65 / 0.018 & 0.905 / 8.71 / 0.246 & 0.897 / 81.3 / 1.874 \\
SAPS not randomised, ours & 0.906 / 1.90 / 0.008 & 0.904 / 8.62 / 0.045 & 0.896 / 81.0 / 0.257 \\
SAPS not randomised, \pkg{TorchCP} & 0.338 / 0.63 / 0.016 & 0.904 / 8.62 / 0.293 & 0.896 / 81.0 / 1.774 \\ \bottomrule
\end{tabular}
\end{table}

\begin{table}[h]\centering\small
\caption{Time-series methods per step when replaying a stream, in microseconds (\code{speed\_timeseries.py}). References were run on shorter streams than \pkg{pyconformal} ($10^6$ steps; 2{,}000 for EnbPI and SPCI), given in the third column, because several slow down per step as the stream grows; the factors would be larger at equal length. DtACI and AgACI are not output-comparable (random expert choice; different aggregation).}\label{tab:online}
\begin{tabular}{@{}llrrrr@{}}\toprule
Method & Reference & Steps & Reference & Ours & Factor \\ \midrule
Conformal PID (P) & authors' code & 20{,}000 & 10.2 & 0.09 & 112 \\
Conformal PID (P+I) & authors' code & 20{,}000 & 14.0 & 0.13 & 106 \\
ACI & authors' code & 20{,}000 & 114 & 0.40 & 286 \\
DtACI, update step only & authors' R code & 50{,}000 & 41 & 0.62 & 66 \\
AgACI, update step only & R code, 30 experts & 50{,}000 & 20 & 4.4 & 4.5 \\
Non-exchangeable CP & definition recomputed each step & 4{,}000 & 71 & 0.30 & 236 \\
EnbPI & \pkg{MAPIE} 1.6.0 & 300 & 5{,}454 & 11.6 & 400--470 \\
EnbPI, batches of 10 & \pkg{PUNCC} 0.9.3 & 2{,}000 & 281 & 12 & 23 \\
SPCI, refit every step & the paper's loop, another forest & 100 & 29{,}100 & 3{,}300 & 9 \\ \bottomrule
\end{tabular}
\end{table}

\begin{table}[h]\centering\small
\caption{MSTL decomposition in milliseconds (\code{speed\_mstl.py}). At its defaults \pkg{statsforecast} evaluates every smoother at every point; R and \pkg{pyconformal} evaluate at every tenth point of the window (the ``jump'' parameters of the original STL), which accounts for most of the gap. At identical settings \pkg{pyconformal} is 6--7 times faster than \pkg{statsforecast} and about 5 times faster than R, with results equal to R's to $10^{-14}$.}\label{tab:mstl}
\begin{tabular}{@{}lrr@{}}\toprule
Series (seasonal periods) & \pkg{statsforecast} 2.1.1 & \pkg{pyconformal} \\ \midrule
15-minute data, 6 weeks, 4{,}032 points & 3{,}145 & 6 \\
Hourly data, 1 year, 8{,}760 points & 1{,}698 & 12 \\
Hourly data, 11 years, 100{,}000 points & 19{,}300 & 98 \\ \bottomrule
\end{tabular}
\end{table}

\begin{table}[h]\centering\small
\caption{Peak extra memory of one call, in megabytes (\code{memory.py}). Each case runs in a fresh process; inputs are built and compiled kernels warmed first, then the peak-memory counter of the process is reset and read after the call. One run on 26 November 2026.}\label{tab:mem}
\begin{tabular}{@{}llrrrr@{}}\toprule
Call & Against & Calibration / test & Theirs & Ours & Ratio \\ \midrule
CV+ prediction, 10 folds & \pkg{MAPIE} & 2{,}000$^{a}$ / 20{,}000 & 1{,}300 & \textbf{3.2} & 406 \\
 & & 5{,}000$^{a}$ / 20{,}000 & 3{,}057 & \textbf{3.2} & 955 \\
CPS, 90\% intervals & \pkg{crepes} & $10^4$ / $10^5$ & 20.1 & \textbf{3.1} & 6 \\
 & & $10^5$ / $10^6$ & 197 & \textbf{31.5} & 6 \\
CPS, distribution function & \pkg{crepes} & $10^4$ / $10^5$ & 6.9 & \textbf{5.5} & 1.3 \\
 & & $10^5$ / $10^6$ & 55.8 & \textbf{54.4} & 1.0 \\
CPS, CRPS & \pkg{crepes} & 5{,}000 / 5{,}000 & 383 & \textbf{0.4} & 956 \\
Venn--Abers, fit and predict & \pkg{venn-abers} & $10^5$ / $10^6$ & 59.1 & \textbf{38.2} & 1.5 \\
 & & $10^6$ / $10^6$ & 155 & \textbf{111} & 1.4 \\ \bottomrule
\end{tabular}

\smallskip{\footnotesize $^{a}$ Training points.}
\end{table}

\begin{table}[h]\centering\small
\caption{Quantile models compared (\code{quantile\_models.py}): pinball loss of the raw quantiles, and mean interval length after CQR and CIR+ at a 90\% target. Five random splits per dataset, half the rows for training and a quarter each for calibration and test; 100 trees for the forests, 300 rounds for the boosters; 2 cores. Coverage after calibration was between 89.4\% and 90.5\% in every row. The three boosters are called through \code{QuantileBooster}. Best value per dataset in bold.}\label{tab:qmodels}
\begin{tabular}{@{}llrrrrr@{}}\toprule
Dataset & Model & Pinball & CQR & CIR+ & Fit (s) & Predict (s) \\ \midrule
California housing & \pkg{quantile-forest} (Zillow) & 0.129 & 1.87 & 1.43 & 1.5 & 0.14 \\
 & \code{QuantileForest} & 0.124 & 1.70 & 1.41 & \textbf{0.23} & 0.08 \\
 & CatBoost & 0.119 & \textbf{1.36} & 1.38 & 6.4 & \textbf{0.02} \\
 & XGBoost & 0.123 & 1.49 & 1.40 & 9.2 & 0.28 \\
 & LightGBM & \textbf{0.116} & \textbf{1.36} & \textbf{1.34} & 7.1 & 0.98 \\
kin8nm & \pkg{quantile-forest} (Zillow) & 0.048 & 0.573 & 0.532 & 0.65 & 0.07 \\
 & \code{QuantileForest} & 0.045 & 0.541 & 0.513 & \textbf{0.11} & 0.03 \\
 & CatBoost & \textbf{0.035} & \textbf{0.427} & \textbf{0.459} & 4.2 & \textbf{0.01} \\
 & XGBoost & 0.045 & 0.529 & 0.557 & 7.0 & 0.11 \\
 & LightGBM & 0.038 & 0.529 & 0.527 & 5.2 & 0.47 \\
Synthetic, skewed & \pkg{quantile-forest} (Zillow) & 0.455 & 5.02 & 4.70 & 1.5 & 0.14 \\
 & \code{QuantileForest} & 0.454 & 4.98 & 4.62 & \textbf{0.20} & 0.08 \\
 & CatBoost & \textbf{0.452} & 4.77 & 4.68 & 5.5 & \textbf{0.02} \\
 & XGBoost & 0.453 & 4.68 & \textbf{4.48} & 8.0 & 0.28 \\
 & LightGBM & 0.454 & \textbf{4.56} & 4.59 & 6.5 & 0.78 \\
Synthetic, two modes & \pkg{quantile-forest} (Zillow) & 0.824 & 6.74 & 7.61 & 1.8 & 0.14 \\
 & \code{QuantileForest} & 0.821 & 6.67 & 8.08 & \textbf{0.21} & 0.08 \\
 & CatBoost & 0.825 & 6.58 & 7.38 & 5.5 & \textbf{0.02} \\
 & XGBoost & \textbf{0.813} & 6.50 & 6.71 & 7.7 & 0.27 \\
 & LightGBM & 0.825 & \textbf{6.44} & \textbf{6.59} & 6.3 & 0.74 \\ \bottomrule
\end{tabular}
\end{table}

\section{Experimental details}\label{app:details}

\paragraph{Environment.} Table~\ref{tab:env} lists the machine and the controls. Thread counts were left at their defaults for every package, which on this machine means two threads for BLAS and for Numba; nothing was pinned to one thread. All arrays are 64-bit floats. Where a competing package offers a vectorised call for many test rows, that call was used; the one-row-at-a-time measurements are labelled as such.

\begin{table}[h]\centering\small
\caption{Environment and controls.}\label{tab:env}
\begin{tabular}{@{}lp{12.2cm}@{}}\toprule
Machine & 2 virtual cores (Intel Xeon, 2.1 GHz), 7 GB memory, Linux \\
Python & 3.13; Numba 0.68.1, scikit-learn 1.9.1, SciPy 1.17.1; NumPy 2.5.5 (OpenBLAS 0.3.34), or NumPy 1.26.4 with PyTorch 2.15 in the environment used for \pkg{MAPIE}, \pkg{TorchCP} and \pkg{PUNCC} \\
Packages & \pkg{MAPIE} 1.6.0, \pkg{crepes} 0.9.1, \pkg{TorchCP} 1.2.1, \pkg{PUNCC} 0.9.3, \pkg{venn-abers} 1.5.4, \pkg{statsforecast} 2.1.1, \pkg{mlforecast} 1.1.0, \pkg{Darts} 0.48.1, \pkg{conformalforecast} 0.1.0 (Nixtla), \pkg{conformal-tights} 0.5.0, \pkg{online-cp} 0.4.0, \pkg{quantile-forest} 1.5.0, R \pkg{forecast} 9.0.2, R \pkg{conformalForecast} 0.2.1.9000 (development version) \\
Threads & defaults (2 for BLAS and Numba); not pinned \\
Precision & 64-bit floats throughout \\
Timing & best of 3 runs (single run where one run exceeds a minute); compiled kernels warmed up first \\
Seeds & fixed in every script (NumPy \code{default\_rng} with the seed given in the script) \\
Quantile forest & measured separately on a 16-core server under load from another job \\ \bottomrule
\end{tabular}
\end{table}

\paragraph{Data.} All data are synthetic, except the three Monash series of Section~\ref{sec:real} and the eight public regression datasets used in the audit of \pkg{conformal-tights}.
\begin{itemize}
\item \emph{Threshold exactness (Table~\ref{tab:exact}).} Calibration scores drawn from an exponential distribution; $n = 1,\dots,120, 199, 200, 999, 1000$ and $\alpha \in \{0.01, 0.02, 0.05, 0.1, 0.15, 0.2, 0.25, 0.3, 0.5, 0.9\}$, 1{,}240 cases. Each package's half-width is compared with the exact order statistic.
\item \emph{\pkg{statsforecast} and \pkg{mlforecast} (Table~\ref{tab:nixtla}).} 4{,}000 series of 120 independent standard normal values; a historic-average forecaster and a linear regression; nominal 90\%.
\item \emph{\pkg{Darts}.} 1{,}000 series of 40 independent standard normal values; linear regression on one lag; calibration lengths 3 to 20.
\item \emph{\pkg{TorchCP} SAPS (Table~\ref{tab:faults}).} 10 classes; SAPS weight 0.2; logits $c\,\mathcal N(0,1)$ with $c \in \{3, 6, 10\}$; labels drawn from the softmax probabilities; 5{,}000 calibration and 50{,}000 test points.
\item \emph{Prediction sets (Table~\ref{tab:sets}).} Logits $1.5\,\mathcal N(0,1)$ plus 4 on the true class; 5{,}000 calibration points; 50{,}000 test points (20{,}000 for 1{,}000 classes).
\item \emph{Venn--Abers (Table~\ref{tab:va}).} Scores uniform on $[0,1]$; labels Bernoulli with probability $s^{1.5}$ (realistic), $1-s$ (anti-correlated) or $0.5$ (noise); or scores rounded to two decimals.
\item \emph{Predictive systems (Table~\ref{tab:cps}).} Standard normal outcomes and predictions; difficulty estimates uniform on $[0.5, 3]$.
\item \emph{Quantile methods.} $y = x\,e$ with $x$ uniform on $[0.5, 2]$ and $e$ gamma with shape 2; 101 predicted quantiles equal to the true ones times a log-normal model error of scale 0.15, plus noise of scale 0.05. Table~\ref{tab:cirabl} also uses a normal and a two-component normal mixture for $e$.
\item \emph{ACI.} A random walk of scale 0.1 plus Student-$t_3$ noise whose scale triples half-way; a linear model on five lags trained once on 250 points; 250 calibration points.
\item \emph{EnbPI.} 1{,}000 training rows, five standard normal features, linear outcome with unit noise, 30 bootstrap models.
\item \emph{Multi-step methods.} An AR(1) series with coefficient 0.8 and its optimal forecasts (multi-step split); an ARMA(1,1) series with Student-$t_5$ innovations and a deliberately misspecified forecaster (AcMCP).
\end{itemize}

\section{Additional experiments}\label{app:extra}

\begin{table}[h]\centering\small
\caption{Latency of one step, one row at a time, in microseconds (predict, then update). The model is a stub that returns its input and optionally waits a fixed time, so that the remaining time is the conformal step. 3{,}000 steps for ACI, 500 for EnbPI.}\label{tab:live}
\begin{tabular}{@{}llrrrr@{}}\toprule
Method, model & Package & Median & 95th pct. & 99th pct. & Max \\ \midrule
ACI, zero-cost model & \pkg{MAPIE} & 620 & 943 & 1{,}054 & 3{,}641 \\
 & \pkg{pyconformal} & 9.3 & 10.9 & 28.9 & 164 \\
ACI, model costing 100 $\mu$s per call & \pkg{MAPIE} & 695 & 902 & 1{,}228 & 2{,}944 \\
 & \pkg{pyconformal} & 111 & 128 & 156 & 575 \\
EnbPI, 30 linear models & \pkg{MAPIE} & 5{,}737 & 8{,}230 & 10{,}378 & 13{,}871 \\
 & \pkg{pyconformal} & 2{,}080 & 2{,}742 & 4{,}515 & 6{,}434 \\ \bottomrule
\end{tabular}
\end{table}

\begin{table}[h]\centering\small
\caption{Venn--Abers on stress configurations: seconds for calibration on $n$ points plus prediction for $n$ points. A ratio of 10 would be exactly linear; sorting accounts for part of the excess. The output was compared with \pkg{venn-abers} on 3{,}000 calibration points per configuration and was identical in every case.}\label{tab:vastress}
\begin{tabular}{@{}lrrr@{}}\toprule
Configuration & $n=10^5$ & $n=10^6$ & Ratio \\ \midrule
Continuous scores, informative & 0.035 & 0.489 & 14.0 \\
Pure noise & 0.030 & 0.391 & 13.0 \\
Anti-correlated & 0.018 & 0.254 & 14.0 \\
Perfectly separated & 0.013 & 0.237 & 18.1 \\
Labels alternate in score order & 0.016 & 0.244 & 14.8 \\
Blocks of $\sqrt{n}$ equal labels & 0.016 & 0.271 & 17.1 \\
Ten distinct scores & 0.005 & 0.059 & 11.4 \\
All scores equal & 0.002 & 0.019 & 11.3 \\
Geometrically spaced scores & 0.005 & 0.064 & 12.0 \\ \bottomrule
\end{tabular}
\end{table}

\begin{table}[h]\centering\small
\caption{Quantile forest: exact leaves against the approximation that keeps one training sample per leaf. CQR on the forest's 5\% and 95\% quantiles at a 90\% target; 15{,}000 training, 5{,}000 calibration and 10{,}000 test points, heteroscedastic Student-$t_4$ noise, 100 trees, 10 repetitions (standard errors in brackets).}\label{tab:forestapprox}
\resizebox{\textwidth}{!}{\begin{tabular}{@{}lrrrrr@{}}\toprule
Mode & Raw coverage & Coverage after CQR & Length after CQR & Pinball loss & Predict (s) \\ \midrule
Exact & 0.900 & 0.901 (0.002) & 3.705 (0.014) & 0.1379 (0.0008) & 0.144 \\
One sample per leaf & 0.895 & 0.902 (0.001) & 3.802 (0.012) & 0.1402 (0.0008) & 0.074 \\ \bottomrule
\end{tabular}}
\end{table}

The approximation halves prediction time and costs 2.6\% in interval length; coverage after conformal calibration is unchanged, as it must be. It is worth using when prediction time matters more than a few per cent of width.

\begin{table}[h]\centering\small
\caption{Randomised APS with many classes: 2{,}000 calibration and 2{,}000 test rows, calibrate plus predict. Time grows in proportion to the size of the probability matrix. Peak memory is for the whole process, including generating the data.}\label{tab:classes}
\resizebox{\textwidth}{!}{\begin{tabular}{@{}rrrrrr@{}}\toprule
Classes & Seconds & ns per row and class & Coverage & Mean set size & Input / peak memory (MB) \\ \midrule
100 & 0.005 & 26 & 0.898 & 8.5 & 3 / 427 \\
1{,}000 & 0.053 & 27 & 0.905 & 82 & 32 / 434 \\
10{,}000 & 0.669 & 34 & 0.903 & 839 & 320 / 1{,}371 \\
30{,}000 & 2.018 & 34 & 0.904 & 2{,}257 & 960 / 3{,}247 \\ \bottomrule
\end{tabular}}
\end{table}

\begin{table}[h]\centering\footnotesize
\caption{Interval methods on predicted quantiles: coverage and mean length, with standard errors in the last digits in brackets. 500 calibration points, 300 repetitions of 2{,}000 test points. ``CIR, authors' rule'' is \code{window="shortest"}; for CTI the length is the total length of the set. On the bimodal outcome the authors' rule under-covers; we confirmed the same numbers with the authors' own code (0.883 and 0.767).}\label{tab:cirabl}
\resizebox{\textwidth}{!}{\begin{tabular}{@{}llrrrrr@{}}\toprule
Outcome & Target & CQR & CIR (default) & CIR, authors' rule & CIR+ & CTI \\ \midrule
Skewed (gamma) & 90\% & .900(1) / 5.59 & .906(1) / 5.12 & .905(1) / 5.11 & .901(1) / 5.04 & .899(1) / 4.95 \\
 & 80\% & .802(1) / 4.30 & .806(1) / 3.84 & .802(1) / 3.78 & .801(1) / 3.79 & .801(1) / 3.86 \\
Symmetric (normal) & 90\% & .900(1) / 4.23 & .904(1) / 4.35 & .902(1) / 4.31 & .900(1) / 4.29 & .900(1) / 4.26 \\
 & 80\% & .799(1) / 3.26 & .804(1) / 3.36 & .800(1) / 3.29 & .799(1) / 3.33 & .800(1) / 3.39 \\
Bimodal (two normals) & 90\% & .900(1) / 7.00 & .906(1) / 7.60 & \textbf{.883(1)} / 6.95 & .901(1) / 7.57 & .900(1) / 4.85 \\
 & 80\% & .801(1) / 6.29 & .806(1) / 7.10 & \textbf{.769(1)} / 6.16 & .802(1) / 7.08 & .800(1) / 3.81 \\ \bottomrule
\end{tabular}}
\end{table}

Table~\ref{tab:cirabl} also shows where each method pays off. On the skewed outcome CIR, CIR+ and CTI are 8--12\% shorter than CQR. On the symmetric outcome CQR is as short as any, and CIR is 2--3\% longer. On the bimodal outcome any single interval must span the gap between the modes, and CTI, which may return a union of intervals, is 30--40\% shorter than the rest.

\end{document}